\documentclass[sn-mathphys,Numbered]{sn-jnl}

\usepackage{graphicx}
\usepackage{multirow}
\usepackage{amsmath,amssymb,amsfonts}
\usepackage{amsthm}
\usepackage{mathrsfs}
\usepackage[title]{appendix}
\usepackage{textcomp}
\usepackage{manyfoot}
\usepackage{booktabs}
\usepackage{algorithm}
\usepackage{algorithmicx}
\usepackage{algpseudocode}
\usepackage{listings}

\usepackage{adjustbox}
\usepackage{rotating}

\usepackage{geometry}
\usepackage{amssymb}
\usepackage{latexsym}
\usepackage{mdframed}
\usepackage[table]{xcolor}
\definecolor{mycolor1}{RGB}{255,255,255}
\definecolor{mycolor2}{RGB}{220,235,237}
\definecolor{mycolor3}{RGB}{0,99,110}

\usepackage{lettrine}

\usepackage{makecell}

\usepackage{caption}
\newlength{\myfontsize}
\newlength{\myrowspacing}
\usepackage{booktabs}

\usepackage{amsmath}
\definecolor{newcolor}{rgb}{.8,.349,.1}

\theoremstyle{thmstyleone}

\theoremstyle{thmstyletwo}

\theoremstyle{thmstylethree}

\begin{document}

\title{Artificial Intelligence for Detecting Fetal Orofacial Clefts and Advancing Medical Education}

\author[1,3,4,9]{\fnm{Yuanji} \sur{Zhang}}
\equalcont{These authors contributed equally to this work.}
\author[1,3,4]{\fnm{Yuhao} \sur{Huang}}
\equalcont{These authors contributed equally to this work.}
\author[5]{\fnm{Haoran} \sur{Dou}}
\equalcont{These authors contributed equally to this work.}
\author[6]{\fnm{Xiliang} \sur{Zhu}}
\equalcont{These authors contributed equally to this work.}
\author[7]{\fnm{Chen} \sur{Ling}}
\author[7]{\fnm{Zhong} \sur{Yang}}
\author[8]{\fnm{Lianying} \sur{Liang}}
\author[9]{\fnm{Jiuping} \sur{Li}}
\author[1,3,4]{\fnm{Siying} \sur{Liang}}
\author[10]{\fnm{Rui} \sur{Li}}
\author[10]{\fnm{Yan} \sur{Cao}}
\author[1,3,4]{\fnm{Yuhan} \sur{Zhang}}
\author[6]{\fnm{Jiewei} \sur{Lai}}
\author[1,3,4]{\fnm{Yongsong} \sur{Zhou}}
\author[11]{\fnm{Hongyu} \sur{Zheng}}
\author[12]{\fnm{Xinru} \sur{Gao}}
\author[13]{\fnm{Cheng} \sur{Yu}}
\author[14]{\fnm{Liling} \sur{Shi}}
\author[1,3,4]{\fnm{Mengqin} \sur{Yuan}}
\author[10]{\fnm{Honglong} \sur{Li}}
\author[1,3,4]{\fnm{Xiaoqiong} \sur{Huang}}
\author[1,3,4]{\fnm{Chaoyu} \sur{Chen}}
\author[1,3,4]{\fnm{Jialin} \sur{Zhang}}
\author[6]{\fnm{Wenxiong} \sur{Pan}}
\author[5,15,16,17]
{Alejandro F. Frangi}
\author*[8]{\fnm{Guangzhi} \sur{He}}\email{810080705@qq.com}
\author*[1,3,4]{\fnm{Xin} \sur{Yang}}\email{xinyang@szu.edu.cn}
\author*[9]{\fnm{Yi} \sur{Xiong}}\email{13352995536@163.com}
\author*[7]{\fnm{Linliang} \sur{Yin}}\email{yllsznthello@hotmail.com}
\author*[7]{\fnm{Xuedong} \sur{Deng}}\email{xuedongdeng@163.com}
\author*[2,3,4,6]{\fnm{Dong} \sur{Ni}}\email{nidong@szu.edu.cn}

\affil[1]{\small National-Regional Key Technology Engineering Laboratory for Medical Ultrasound, School of Biomedical Engineering, Shenzhen University Medical School, Shenzhen University, Shenzhen 518060, China}
\affil[2]{\small National Engineering Laboratory for Big Data System Computing Technology, Shenzhen University, Shenzhen 518060, China}
\affil[3]{\small Medical Ultrasound Image Computing (MUSIC) Laboratory, Shenzhen University, Shenzhen 518060, China}
\affil[4]{\small Marshall Laboratory of Biomedical Engineering, Shenzhen University, Shenzhen 518060, China}
\affil[5]{\small Christabel Pankhurst Institute, Department of Computer Science, School of Engineering, University of Manchester, Manchester M13 9PL, UK}
\affil[6]{\small School of Biomedical Engineering and Informatics, Nanjing Medical University, Nanjing 211166, China}
\affil[7]{\small The Affiliated Suzhou Hospital of Nanjing Medical University, Suzhou 215002, China}
\affil[8]{\small Shenzhen University of Advanced Technology General Hospital, Shenzhen 518107, China}
\affil[9]{\small Shenzhen Luohu People’s Hospital, Shenzhen 518001, China}
\affil[10]{\small Shenzhen RayShape Medical Technology Co., Ltd, Shenzhen 518060, China}
\affil[11]{\small The People’s Hospital of Guangxi Zhuang Autonomous Region, Nanning 530021, China}
\affil[12]{\small Northwest Women’s and Children’s Hospital, Xi'an 710061, China}
\affil[13]{\small Hangzhou Women’s Hospital, Hangzhou 310006, China}
\affil[14]{\small Shanxi Children’s Hospital, Taiyuan 030001, China}
\affil[15]{\small Christabel Pankhurst Institute, Division of Informatics, Imaging \& Data Sciences, University of Manchester, Manchester M13 9PL, UK}
\affil[16]{\small NIHR Manchester Biomedical Research Centre, Manchester Academic Health Sciences Centre, University of Manchester, Manchester M13 9PL, UK}
\affil[17]{\small Medical Imaging Research Centre (MIRC), Department of Cardiovascular Sciences and Department of Electrical Engineering, KU Leuven 3000, Leuven, Belgium}

\abstract{

Orofacial clefts are among the most common congenital craniofacial abnormalities, yet accurate prenatal detection remains challenging due to the scarcity of experienced specialists and the relative rarity of the condition. Early and reliable diagnosis is essential to enable timely clinical intervention and reduce associated morbidity. Here we show that an artificial intelligence system, trained on over 45,139 ultrasound images from 9,215 fetuses across 22 hospitals, can diagnose fetal orofacial clefts with sensitivity and specificity exceeding 93$\%$ and 95$\%$ respectively, matching the performance of senior radiologists and substantially outperforming junior radiologists. When used as a medical copilot, the system raises junior radiologists' sensitivity by more than 6 $\%$. Beyond direct diagnostic assistance, the system also accelerates the development of clinical expertise. A pilot study involving 24 radiologists and trainees demonstrated that the model can improve the expertise development for rare conditions. This dual-purpose approach offers a scalable solution for improving both diagnostic accuracy and specialist training in settings where experienced radiologists are scarce.

}
\maketitle

\section{Introduction}\label{sec1} 

Orofacial cleft (OC) is one of the most common craniofacial congenital anomalies, affecting approximately 1 in 700 live births globally~\cite{Nyberg1995, maarse2010diagnostic,Gillham2009}, 30\% of which are syndromic, occurring with other anomalies, such as cardiac and craniospinal defects or chromosomal problems, etc~\cite{Berg2001, kraus1963malformations,benacerraf2019paramedian}. Early and accurate OC diagnosis enables optimal management, improving outcomes of craniofacial disabilities and reducing mortality risks~\cite{Rotten2004, pilu1986prenatal, purisch2008preterm,ko2020multidisciplinary}. Despite advances in prenatal imaging, accurate diagnosis remains challenging, particularly in early gestation where anatomical features are subtle, and less experienced radiologists often struggle to identify these nuanced presentations~\cite{yu2023performance,de2019cleft,platt2006improving}. This diagnostic challenge raises a two-fold problem in modern healthcare: the immediate need for accurate, timely detection and the long-term necessity of developing clinical expertise in this specialized field.

Current prenatal diagnosis of OC relies heavily on subjective interpretation of ultrasound images, with accuracy rates varying significantly based on radiologists' experience (from 9\% to 88\% as reported in~\cite{Johnson2000,Demircioglu2008,Offerdal2008,weissbach2024hard,Chaoui2015}). This expertise variation not only impacts immediate patient care but also raises substantial challenges in training new specialists because the clinical practice struggles to provide sufficient exposure to these relatively rare diseases~\cite{maarse2010diagnostic,Salomon2022,Johnson2000,Demircioglu2008,Offerdal2008,weissbach2024hard,Chaoui2015}. Moreover, anatomical complexity of OC and their rapid changes with gestational age exacerbated the difficulty of developing expertise~\cite{Nicot2019,tian2019accurate,Ji2021,Sommerlad2010,Liou2011}. Typically, cultivating to expert level with clinical pertinent and technical accuracy can take 5-10 years, which makes it hard to catch the growing global demand for prenatal diagnosis, raising an urgent need for a scalable solution that can both augment diagnostic capabilities and accelerate expertise development.

Recent advancements in artificial intelligence (AI) algorithms, particularly deep learning (DL), have revealed the potential to provide expert-level diagnosis in prenatal routine, with applications in fetal structure identification, biometry measurement, and malformations diagnosis (e.g., neurological or cardiac diseases)~\cite{ramirez2023use,smak2023artificial,madani2018fast,lin2023deep,vasey2022reporting,platt2006improving}. However,  these solutions have frequently demonstrated reduced performance when applied across different gestational ages or in external validation cohorts, limiting their practicality in diverse clinical settings. Currently, the application of AI to OC diagnosis remains unexplored, primarily due to the inherent complexity of craniofacial structures and the relative scarcity of annotated datasets~\cite{beaty2016genetic,world2022global}. 
Building an AI-based model trained with large-scale, diverse datasets can provide valuable diagnostic assistance in clinical practice, particularly helping to mitigate diagnostic errors associated with limited clinical experience.

A critical yet unaddressed question in this field is whether such developed computer-aided diagnostic tools impact the development of clinical expertise among radiologists. Current diagnostic systems~\cite{hannun2019cardiologist,Wilkes2023} predominantly focus on evaluating the diagnosis accuracy while overlooking their potential educational utility. For rare diseases, such as OC, where training opportunities are inherently limited, specialized AI methodologies that provide explainable outputs could potentially serve dual purposes, improving immediate diagnostic accuracy while simultaneously enhancing radiologists' ability to identify critical anatomical features through continuous exposure and feedback.

\begin{figure}[!t]
    \centering
    \includegraphics[width=1\textwidth,height=\textheight,keepaspectratio]{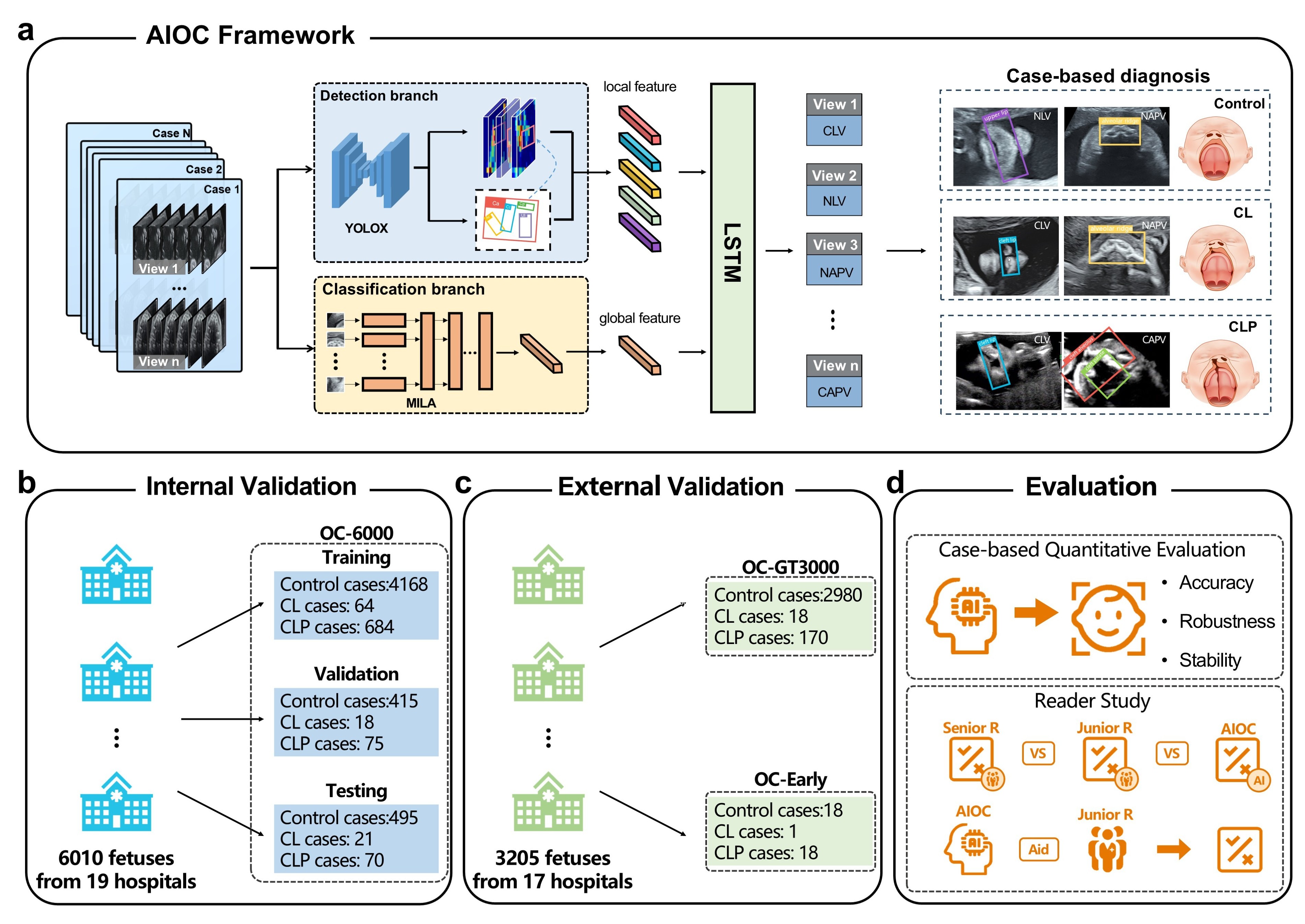}
    \caption{\textbf{Design of the AIOC system.} (a) Development of the AIOC system. 
    It is based on a dual-branch network, integrates detection (YOLOX) and classification (MILA) branches for accurate diagnosis of OC and healthy cases (i.e., CL, CLP, and Control). It also visualizes view classification (i.e., NLV, CLV, NAPV, CAPV) and key structures (i.e., upper lip (purple), alveolar ridge (yellow), cleft lip (blue), cleft alveolus (red), and cleft palate (green)). 
    (b) The internal dataset, OC-6000. (c) The external datasets, OC-GT3000 and OC-Early. (d) Case-based quantitative evaluation of AIOC and a comparative experiment with radiologists. CL: cleft lip; CLP: cleft lip and palate; CLV: cleft lip view; NLV: normal lip view; NAPV: normal alveolus and palate view; CAPV: cleft alveolus and palate view.}
    \label{fig: overview}
\end{figure}

In this study, we aimed to develop an AI-assisted diagnostic system for orofacial cleft (AIOC) that achieves expert-level performance while concurrently facilitating radiological expertise development.
To support the model development, we first constructed a large-scale, multi-center dataset that consisted of over 40,000 fetal ultrasound images from 9215 subjects with 1,139 OC fetuses and 8,076 healthy fetuses. We then developed, validated, and externally tested our AIOC system in this collected dataset to comprehensively evaluate its performance in OC diagnosis. Results demonstrated the superiority of the proposed AIOC system in OC diagnosis with respect to accuracy, stability, and robustness. We further constructed a pilot study for the AIOC system to investigate its potential to improve the long-term outcome of training clinical experts for OC analysis. To the best of our knowledge, this is the first study to investigate both diagnostic accuracy and the educational utility of the computer-assisted tool.

\begin{figure}[!t]
    \centering
    \includegraphics[width=1\textwidth,height=\textheight,keepaspectratio]{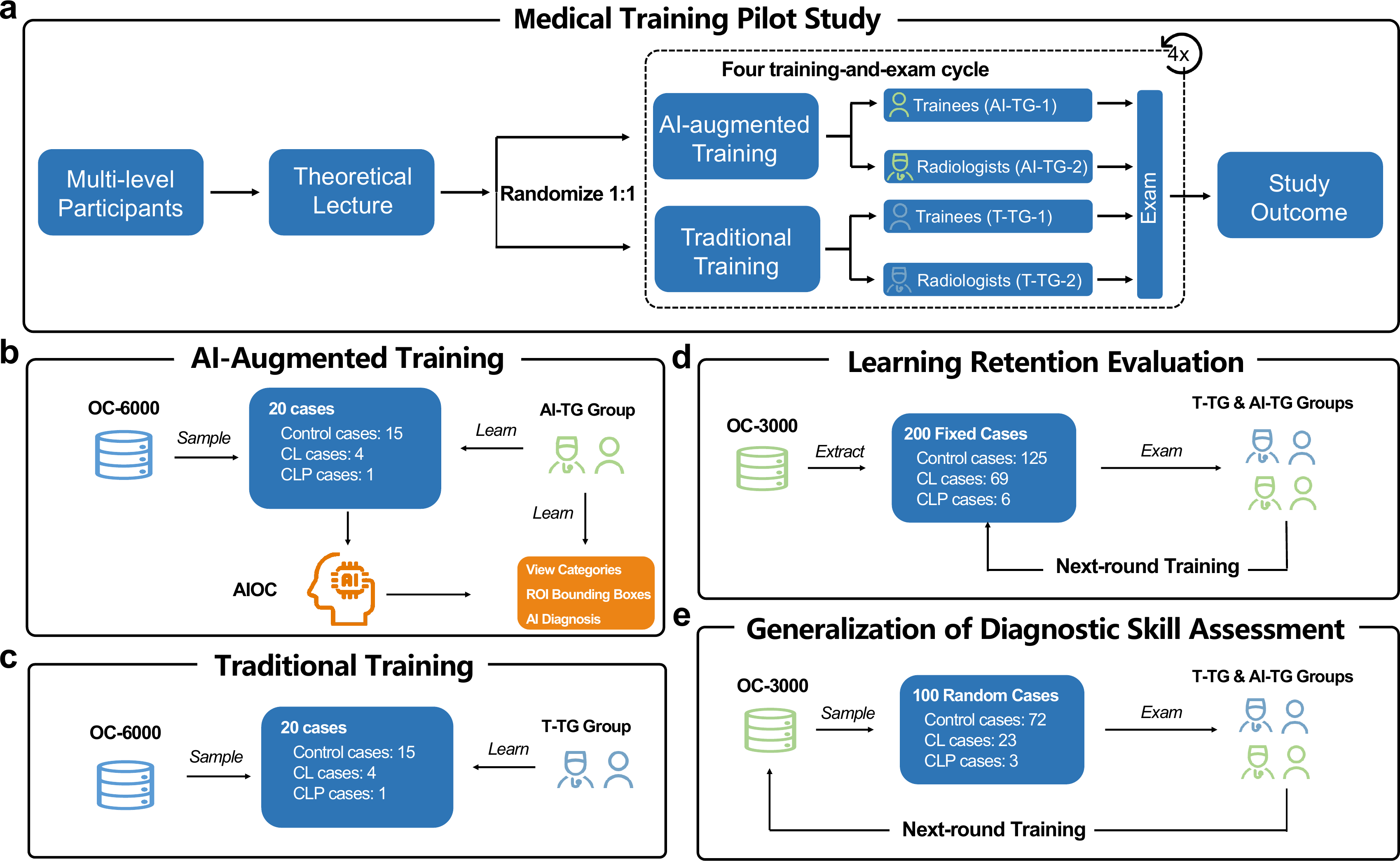}
    \caption{\textbf{Medical training pilot study design.} (a) Overview of the pilot study workflow with theoretical lecture, randomized allocation to AI-augmented or traditional training groups, and four training-and-exam cycles. (b) AI-augmented training using 20 cases from OC-6000 with AIOC system assistance. The participants will be provided with additional view categories, bounding boxes for anatomical structures, and AI diagnostic recommendations, compared with the traditional training group (c). Traditional training using the same 20 cases without AI assistance. (d) Learning retention evaluation using 200 fixed cases from OC-3000. (e) Generalization assessment using 100 random cases from OC-3000 to test diagnostic skill transferability. T-TG: traditional training groups; AI-TG: AI training groups; CL: cleft lip; CLP: cleft lip and palate.}
    \label{fig: education}
\end{figure}

\section{Results}\label{sec2}
\subsection{Study design}
\label{sec:study_design}

To train the AIOC system (Fig. \ref{fig: overview}(a)). and to conduct a thorough validation, this study retrospectively gathered a large-scale dataset comprising 9,215 pregnant women (age: 29.3$\pm$6.8 years) with singleton fetuses, spanning gestational weeks 14 to 28, from 22 hospitals (see Supplementary Table~\ref{tab: S2} for further details). The recruited dataset comprised 45,139 fetal ultrasound images, including 7,559 images from 1,139 cases of OC fetuses and 37,580 images from 8,076 cases of healthy fetuses. Each case contained between 2 and 26 views, with an average of 5 views. The detailed demographics of pregnant women and fetuses, and the ultrasound equipment used, can be found in Table~\ref{tab: S1}.
The AIOC system was trained and validated internally using a subset of the collected datasets, designated as OC-6000 (Fig. \ref{fig: overview}(b)). This dataset contains 28,994 images from 6,010 fetuses spanning gestational weeks 18 to 28. It was divided at the case level into training, validation, and testing sets based on the proportions of 80$\%$, 10$\%$, and 10$\%$, respectively. The proportions of OC cases in each subset are 15.2$\%$, 18.3$\%$, and 15.5$\%$. To further evaluate the AIOC system’s stability and generalizability, the remaining dataset was split into two parts with varying gestational ages (GA) for external validation, OC-GT3000 and OC-Early (Fig. \ref{fig: overview}(c)). The OC-GT3000 dataset, consisting of 15,848 images from 3,168 fetuses within GA 18–28 weeks, mirrors real-world disease incidence rates; the OC-Early dataset, comprising 297 images from 37 fetuses within GA 14-18 weeks, was employed to assess the model's stability during the early gestational stages.

We annotated the dataset with expert input, including three types of diagnoses, i.e., cleft lip (CL), cleft lip and palate (CLP), and healthy fetuses (Control), four view classifications, i.e., Normal Lip View (NLV), Normal Alveolus and Palate View (NAPV), Cleft Lip View (CLV), and Cleft Alveolus and Palate View (CAPV), and five key structures (the upper lip, alveolar ridge, cleft lip, cleft alveolus, and cleft palate). The AIOC system follows a case-based diagnosis procedure for identifying orofacial cleft types by using specific views and targeted structures: CL cases are diagnosed using CLV and NAPV views with structures like the cleft lip and normal alveolar ridge; CLP cases are identified through CLV and CAPV views with structures such as the cleft lip, cleft alveolus, and cleft palate; control cases are recognized using NLV and NAPV views, featuring structures like the upper lip and alveolar ridge (Fig. \ref{fig: overview}(a)).

To evaluate the diagnostic performance of the AIOC system and its effectiveness in assisting junior radiologists, we conducted a reader study to compare the performance of AIOC, junior, senior, and AI-assisted junior radiologists on the OC-GT3000 dataset. Six radiologists participated in the reader study, comprising three senior radiologists (R1-R3, over 10 years of experience) and three junior radiologists (R4-R6, 3 years of experience). Majority voting across the junior/senior radiologists was used to obtain the consensus diagnosis.

To investigate the educational capacity of AIOC in diagnostic assistance, we conducted a four-cycle training-and-exam study, as shown in Fig.~\ref{fig: education} (a). This study recruited 24 participants: 12 trainees (0-1 year experience) and 12 junior radiologists (1-3 years experience), randomized into Traditional Training Groups (T-TG1, T-TG2) or AI Training Groups (AI-TG1, AI-TG2). Each participant completed four weekly training-and-exam cycles. Each cycle consisted of: (1) a training phase using 20 cases from the OC-6000 dataset, and (2) an examination with 300 cases from the OC-GT3000 dataset - comprising 200 fixed cases that remained consistent across all cycles, and 100 novel cases per round. Performance on fixed cases assessed learning retention, while performance on novel cases evaluated generalization of diagnostic skills (Fig. \ref{fig: education}(d) and (e)).

\begin{figure}[!t]
    \centering
    \includegraphics[width=1\textwidth,height=\textheight,keepaspectratio]{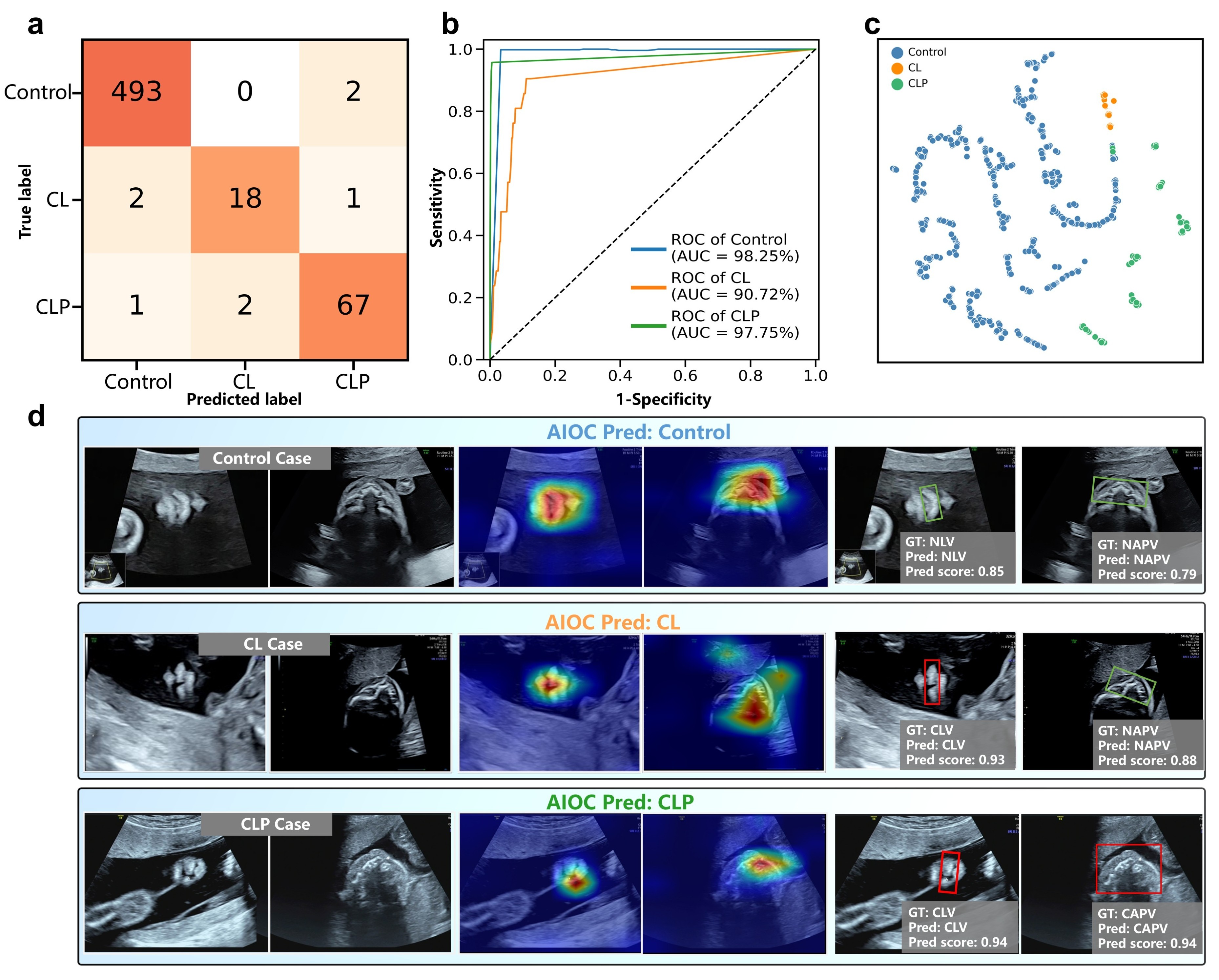}
    \caption{\textbf{Performance of the AIOC system on the OC-6000 testing dataset.} (a) Confusion matrix, (b) ROC curve, and (c) t-SNE plot for case-based diagnosis (Control (blue), CL (orange), and CLP (green)). 
    (d) Specific examples of Control, CL, and CLP cases, comparing the views, Gray-CAM results of AIOC, and the annotated key structures.
    Red boxes highlight cleft lip, cleft alveolus, or cleft palate abnormalities, while green boxes indicate normal tissue regions in control cases. CL: cleft lip; CLP: cleft lip and palate; CLV: cleft lip view; NLV: normal lip view; NAPV: normal alveolus and palate view; CAPV: cleft alveolus and palate view; AUC: area under the curve; ROC: receiver operating characteristic.}
    \label{fig: diagnosis_oc6000}
\end{figure}

\subsection{Performance on case-based diagnoses}
To evaluate the diagnoses performance of the AIOC system, we trained and validated the model internally on the OC-6000 dataset. 
The AIOC system demonstrated superior performance, achieving an average AUC of 95.57$\%$, an F1 score of 94.34$\%$, a sensitivity of 93.67$\%$, a specificity of 98.59$\%$, and an accuracy of 99.09$\%$. Additionally, the false negative rate (FNR) and false positive rate (FPR) were notably low, at 6.33$\%$ and 1.41$\%$, respectively (Table \ref{tab:diagnostic_performance-oc3000}). The confusion matrix (Fig.~\ref{fig: diagnosis_oc6000}(a)) reveals the strong classification performance of the AIOC system across all three categories. ROC curve analysis (Fig.~\ref{fig: diagnosis_oc6000}(b)) confirms the model's superior discriminative ability, with high AUC values for all three categories (Control: 98.25$\%$, CL: 90.72$\%$, CLP: 97.75$\%$), indicating reliable diagnostic performance across all conditions. The t-SNE visualization (Fig. \ref{fig: diagnosis_oc6000}(c)) demonstrates clear clustering patterns of the extracted features, with Control samples (blue) forming distinct clusters separate from pathological cases, while CL (orange) and CLP (green) samples form their own identifiable clusters, further supporting the AIOC system can distinguish between these diagnostic categories.

\begin{figure}[!t]
    \centering
    \includegraphics[width=1\textwidth,height=\textheight,keepaspectratio]{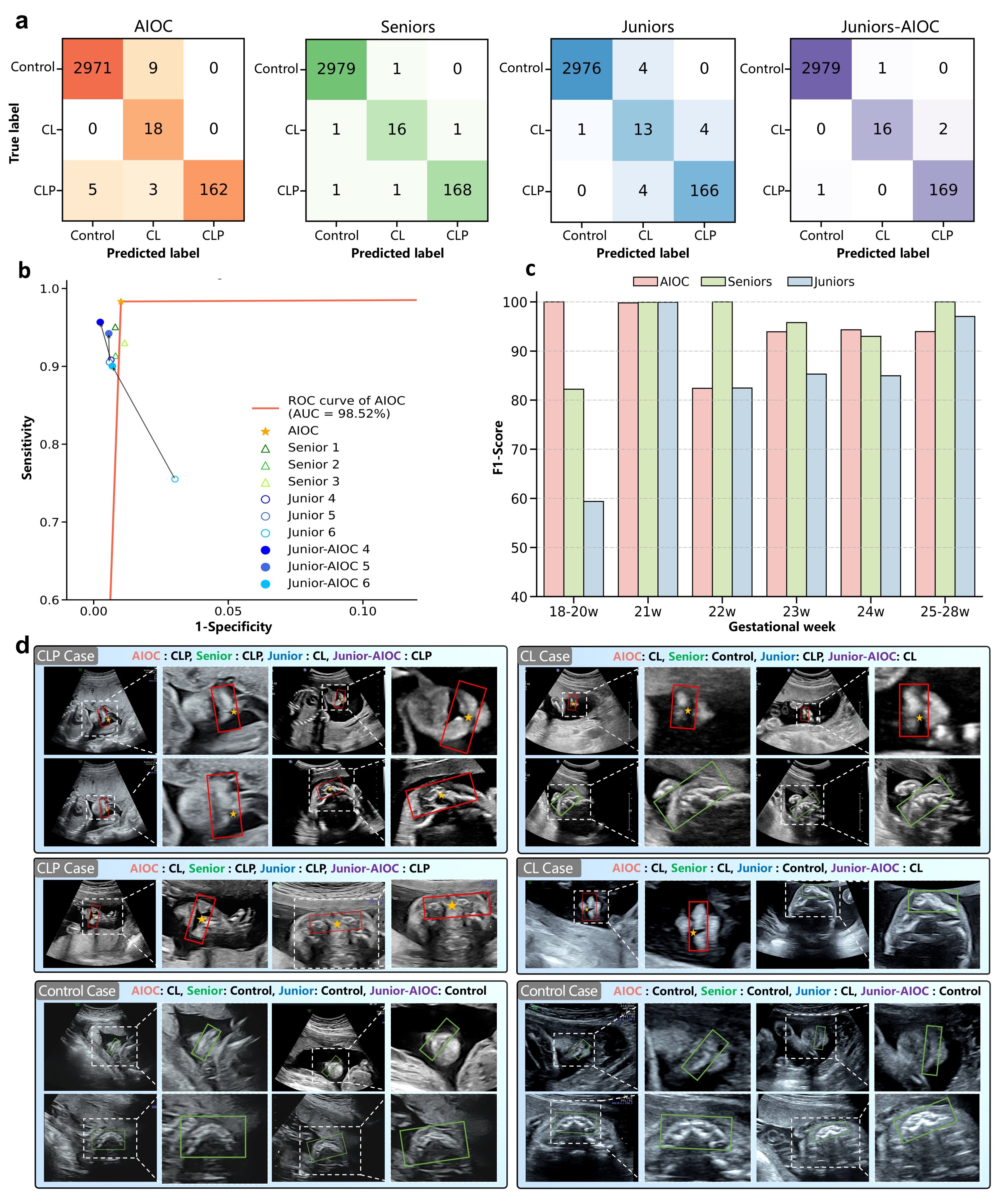}
    \caption{\textbf{Fetal case-based diagnosis results of the AIOC, senior\&junior radiologists, and junior radiologists assisted by AIOC (Junior-AIOC) on the OC-GT3000 dataset.} 
    (a) Confusion matrix and (b) ROC curve for AIOC, Senior, Junior, and Junior-AIOC diagnosis. 
    (c) Bar plot showing F1 scores (18-28 weeks) for AIOC and radiologists (Senior\&Junior). 
    Due to the small number of cases in the 18-20 and 25-28 week groups, they were combined for analysis. 
    (d)
    Representative examples of CLP, CL, and Control cases and their diagnosed results by AIOC, junior radiologists, senior radiologists, and junior radiologists assisted by AIOC. Each case includes multiple ultrasound views (left) and their zoom-in views of the region of interest (right). Red boxes highlight cleft lip, cleft alveolus, or cleft palate abnormalities, while green boxes indicate normal tissue regions in control cases. CL: cleft lip; CLP: cleft lip and palate; AUC: area under the curve; ROC: receiver operating characteristic.
    }
    \label{fig: diagnosis_oc3000}
\end{figure}

To assess the generalizability of the AIOC system across different centers with varying data collection protocols, we validated our model on the external dataset OC-GT3000. 
Table~\ref{tab:diagnostic_performance-oc3000} shows that the AIOC system exhibited robust performance on the OC-GT3000 dataset, 
achieving an average AUC of 98.52$\%$, an F1 score of 90.79$\%$, a sensitivity of 98.33$\%$, a specificity of 98.99$\%$, and an overall accuracy of 99.64$\%$. Meanwhile, FNR and FPR were relatively low at 1.67$\%$ and 1.01$\%$, respectively. The confusion matrices (Fig.~\ref{fig: diagnosis_oc3000}(a)) and ROC curve (Fig.~\ref{fig: diagnosis_oc3000}(b)) of the OC-GT3000 dataset further elucidate the AIOC system’s diagnostic performance in classifying these diagnostic categories.

\begin{figure}[!t]
    \centering
    \includegraphics[width=1\textwidth,height=\textheight,keepaspectratio]{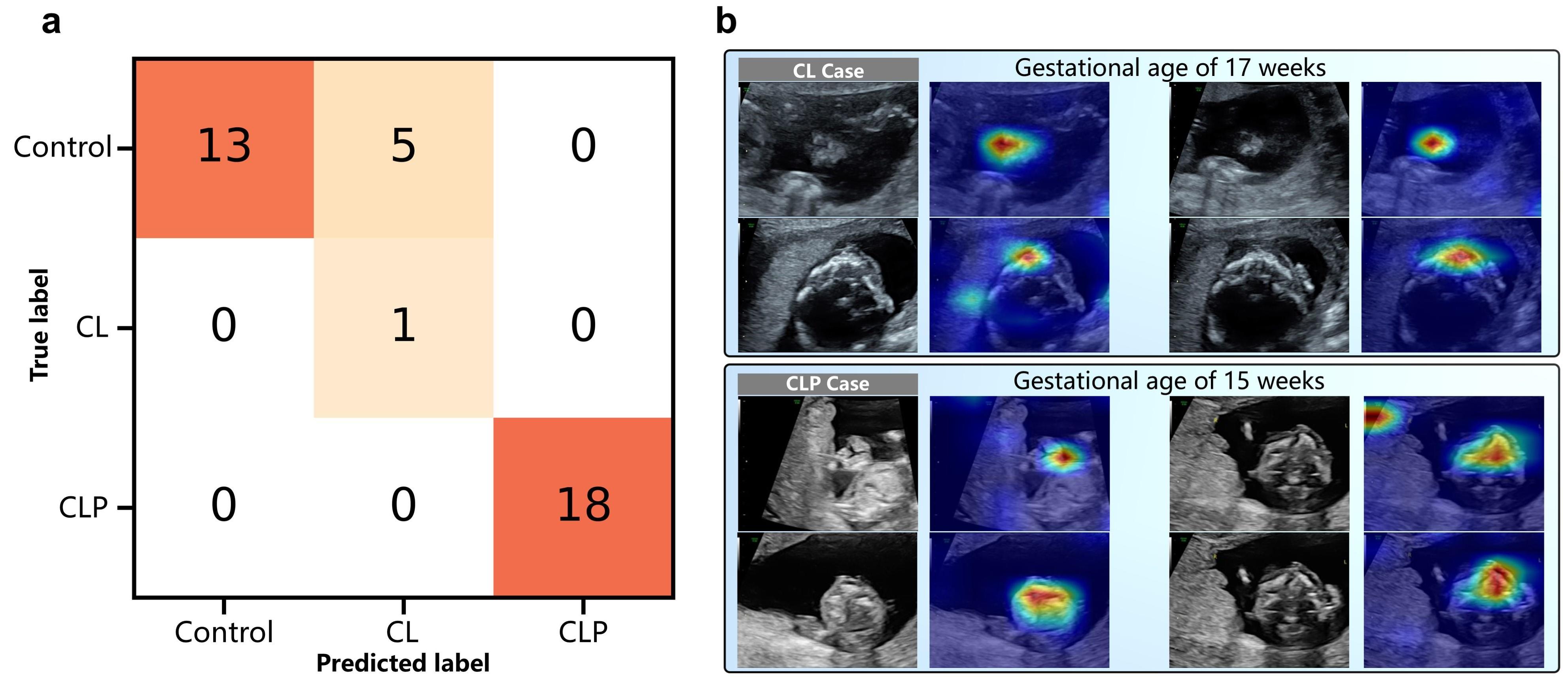}
    \caption{\textbf{Fetal case-based diagnosis results from the AIOC system on the OC-Early dataset.} (a) Confusion matrix for fetal case-based diagnosis. (b) Specific cases: A CL case at 17 weeks of gestational age, with AIOC providing the correct diagnosis and Gray-CAM visualizing the diagnostic details. A CLP case at 15 weeks of gestational age, with AIOC also providing the correct diagnosis and visualizing the details. CL: cleft lip; CLP: cleft lip and palate.}
    \label{fig: diagnosis_early}
\end{figure}

We validated our model on the fetal cases in the early gestation with a focus on 14 to 17 weeks (OC-Early datasets). 
This evaluation explores the feasibility of the developed AIOC system for early-trimester diagnoses. 
The model showed reasonable performance at early gestational OC diagnoses despite lacking specific training data for 14-17 weeks (Table \ref{tab:diagnostic_performance-oc3000}). It achieved an AUC of 93.06\%, an F1 score of 70.81\%, a sensitivity of 90.74\%, a specificity of 95.37\%, and a relatively low FNR (9.26\%) and FPR (4.63\%). The confusion matrix of OC-Early (Fig.~\ref{fig: diagnosis_early}(a)) demonstrates that the AIOC system can provide diagnostic support for early detection, though with some limitations. Most misclassifications occurred between control and CL cases, likely reflecting the increased difficulty of distinguishing subtle anatomical features at earlier gestational ages.

To further investigate the explainability of the AIOC system, we performed gradient-weighted class activation mapping (Grad-CAM) on orofacial images. 
The Grad-CAM results indicated that the AIOC system can focus on clinically relevant features for both OC and healthy cases during the second trimester in OC-6000 (Fig. \ref{fig: diagnosis_oc6000}(d)) and early second trimester in OC-Early (Fig. \ref{fig: diagnosis_early}(b)). 
Specifically, the provided three CAM examples in Fig. \ref{fig: diagnosis_oc6000}(d)  (the middle two images) indicate that the model can capture the key structural regions that match the expert knowledge (refer to the bounding box shown in the last two columns), thus making correct decisions. Detailed analysis of misclassified cases and attention patterns in challenging scenarios is presented in Section~\ref{sec3} and Figure~S2. The four-view classification results on OC-6000 and OC-GT3000 datasets are presented in Supplementary Table~\ref{tab: S3} and Extended Data Fig. \ref{fig: S1}, while the structure detection capabilities are detailed in Supplementary Table~\ref{tab: S4}.

\begin{table}[!t]
\centering
\caption{\textbf{Diagnostic performance of AIOC on the OC-6000, OC-GT3000, and OC-Early datasets, and comparison with senior, junior, and AIOC-assisted junior (Junior-AIOC) radiologists on the OC-GT3000 dataset.} 
The \textbf{OC-6000} presents AIOC's fetal OC case-based diagnostic performance across Control, CL, and CLP categories. 
The "AIOC" rows report the average performance metrics across these categories. 
The  \textbf{OC-GT3000} compares AIOC with Senior, Junior, and Junior-AIOC radiologists. 
They represent the majority voting across three senior, junior, and Junior-AIOC radiologists, respectively. 
The  \textbf{OC-Early} shows AIOC's performance in the early stage (14-17 weeks gestational age). 
AUC: area under the receiver operating characteristic curve; CI: confidence interval; FPR: false positive rate; FNR: false negative rate; F1: F1-score.}
\label{tab:diagnostic_performance-oc3000}
\begin{tabular}{cccccccccc}
\toprule
Characteristic & AUC (\%) & Sensitivity (\%) & Specificity (\%) & Accuracy & FNR & FPR & F1 & Youden \\
 & (95\% CI) & (95\% CI) & (95\% CI) & (\%) & (\%) & (\%) & (\%) & Index \\
\midrule
\textbf{OC-6000} &  &  &  &  &  &  &  &  \\
Control & 98.25 & 99.60 & 96.70 & 99.15 & 0.40 & 3.30 & 99.50 & 0.96 & \\
 & (96.05 - 99.80) & (98.55 - 99.95) & (90.67 - 99.31) &  &  &  &  &  &\\
CL & 90.72 & 85.71 & 99.65 & 99.15 & 14.29 & 0.35 & 87.80 & 0.85 &  \\
 & (84.44 - 99.82) & (70.75 - 100.00) & (99.16 - 100.00) &  &  &  &  &  &  \\
CLP & 97.75 & 95.71 & 99.42 & 98.98 & 4.29 & 0.58 & 95.71 & 0.95 &  \\
 & (94.91 - 99.72) & (90.97 - 100.00) & (98.76 - 100.00) &  &  &  &  &  &  \\
AIOC & 95.57 & 93.67 & 98.59 & 99.09 & 6.33 & 1.41 & 94.34 & 0.92 &  \\
 & (91.80 - 99.78) & (87.04 - 100.00) & (96.98 - 100.00) &  &  &  &  &  &  \\
\hline
\textbf{OC-GT3000} &  &  &  &  &  &  &  &  \\
AIOC & 98.52 & 98.33 & 98.99 & 99.64 & 1.67 & 1.01 & 90.79 & 0.97 &  \\
 & (97.27 - 99.60) & (90.61 - 99.27) & (97.71 - 99.64) &  &  &  &  &  &  \\
Senior & 97.75 & 95.89 & 99.61 & 99.89 & 4.11 & 0.39 & 95.98 & 0.96 &  \\
& (94.43 - 99.99) & (86.97 - 99.49) & (98.60 - 99.95) &  &  &  &  &  &  \\
Junior & 94.80 & 89.91 & 99.69 & 99.73 & 10.09 & 0.31 & 88.08 & 0.90 &  \\
& (90.47 - 98.51) & (80.09 - 96.54) & (98.74 - 99.95) &  &  &  &  &  &  \\
Junior-AIOC& 97.94 & 96.09 & 99.79 & 99.92 & 3.91 & 0.21 & 96.84 & 0.96 &  \\
& (94.61 - 100.00) & (87.29 - 99.54) & (98.88 - 99.99) &  &  &  &  &  &  \\
\hline
\textbf{OC-Early} &  &  &  &  &  &  &  &  \\
AIOC & 93.06 & 90.74 & 95.37 & 90.99 & 9.26 & 4.63 & 70.81 & 0.86 &  \\
& (87.38 - 97.86) & (83.84 - 96.77) & (78.40 - 98.44) &  &  &  &  &  &  \\
\bottomrule
\end{tabular}
\end{table}

\subsection{Comparison with radiologists}
As summarized in Table~\ref{tab:diagnostic_performance-oc3000}, the results of the three senior radiologists exhibit a sensitivity of 95.89$\%$, a specificity of 99.61$\%$, and an AUC of 97.75$\%$, while the three junior radiologists achieved a sensitivity of 89.91$\%$, a specificity of 99.69$\%$, and an AUC of 94.80$\%$. These can be compared with the AIOC system, which achieved a sensitivity of 98.33$\%$, a specificity of 98.99$\%$, and an AUC of 98.52$\%$ on the same dataset. The detailed performance of the six radiologists is provided in Extended Data Table \ref{tab: S5} and Fig. \ref{fig: S2}. 

Confusion matrices (Fig. \ref{fig: diagnosis_oc3000}(a)) and ROC curves  (Fig. \ref{fig: diagnosis_oc3000}(b)) illustrate the performance of the AIOC system, senior radiologists, and junior radiologists in diagnosing  OC and healthy cases. Fig.\ref{fig: diagnosis_oc3000}(d) visualizes representative misclassified cases. The AIOC system's primary error was misclassifying control cases as CL, primarily due to blurred images that obscured the upper lip structure. Senior radiologists maintained low, balanced false-positive and false-negative rates, they occasionally made errors when image quality was suboptimal. Junior radiologists tended to misclassify both CLP and control cases as CL, suggesting insufficient experience with this pathology. The ROC analysis showed that the AIOC system achieved higher sensitivity but with a slightly elevated false positive rate compared to radiologists. Senior radiologists demonstrated the most balanced performance, while junior radiologists had lower sensitivity than both the AIOC system and senior radiologists.

To assess the diagnostic stability of the AIOC system and radiologists across different gestational weeks, we calculated week-specific F1 scores from 18 to 28 weeks. Standard deviation (SD) of the F1 scores across weeks was calculated to estimate the variance in diagnostic performance (Supplementary Table~\ref{tab: S6-Gestational week}). The results consistently demonstrated the AIOC system's stable performance, with F1 scores ranging from 82.40$\%$ to 100.00$\%$ (SD: 5.84). In comparison, senior radiologists show similar performance, with F1 scores from 82.22$\%$ to 100.00$\%$ (SD: 6.35); while junior radiologists demonstrated a broader range from 59.37$\%$ to 99.93$\%$ (SD: 13.11). Senior radiologists and the AIOC system showed similar SD values and F1 score ranges, while junior radiologists' larger SD values and wider F1 score ranges reflected greater diagnostic instability across weeks. The results are depicted in Fig. \ref{fig: diagnosis_oc3000}(c).

Overall, the AIOC system's performance closely matched the accuracy and stability of senior radiologists and exceeded that of junior radiologists in case-based OC diagnosis. The chi-square test was used to compare the diagnostic accuracy between individual radiologists and the AIOC system. For senior radiologists, the analysis yielded \textit{p}-values of 0.45, 0.58, and 0.72 for radiologists R1, R2, and R3, respectively, indicating no statistically significant differences from the AIOC system. For junior radiologists, \textit{p}-values of 0.58 and 0.43 were obtained for radiologists R4 and R5, similarly demonstrating no significant difference from the AIOC system. In contrast, junior radiologist R6 showed a statistically significant difference from the AIOC system (\textless0.001), suggesting comparatively lower diagnostic accuracy relative to the AI system.
Moreover, the AIOC system shows superior efficiency over radiologists, diagnosing in 0.32 seconds compared to 10.54 seconds for seniors and 11.93 seconds for juniors (Supplementary Table~\ref{tab: S6-Time requirement}).

\subsection{Diagnostic performance of AI-Assisted junior radiologists}
To verify the ability of the AIOC system to assist radiologists in improving diagnostic accuracy, three junior radiologists conducted diagnoses assisted by the AIOC system, on the OC-GT3000 dataset. 
We deployed the AIOC system into cloud-based and mobile applications to enable convenient access and to standardize diagnostic support across participating centers (Supplementary Fig.~\ref{fig: S3}). 
The visualization of the bounding box (key structures) and OC diagnostic types was displayed as references to aid junior radiologists with enhanced justification and interpretation. After reviewing clinical cases and the AIOC results, the radiologists made final diagnostic decisions.

The performance of junior radiologists assisted by the AIOC system surpassed their independent diagnosis and was comparable to that of senior radiologists (Table \ref{tab:diagnostic_performance-oc3000}). They achieved a sensitivity of 96.09$\%$, a specificity of 99.79$\%$, and an AUC of 97.94$\%$, outperforming those without assistance by 6.18$\%$, 0.10$\%$, and 3.14$\%$, respectively. The integration of AIOC with junior radiologists (Junior-AIOC) demonstrated expert-level error rates, achieving an FNR of 3.91$\%$ and an FPR of 0.21$\%$, compared to 4.11$\%$~/~0.39$\%$ for senior radiologists. The confusion matrix (Fig. \ref{fig: diagnosis_oc3000}(a)) demonstrates the improvement in case-based diagnostic results for the three OC and healthy types (i.e., CL, CLP, and Control) when junior radiologists are assisted by the AIOC system, particularly in reducing misdiagnosis of CL cases. The ROC curves (Fig. \ref{fig: diagnosis_oc3000}(b)) highlight the sensitivity improvements of the three junior radiologists with AIOC assistance. 
We present one CL and one CLP case (Fig. \ref{fig: diagnosis_oc3000}(d)). In the CL case, AIOC correctly diagnosed it as CL, while the senior radiologist misdiagnosed it as Control and the junior as CLP. With AIOC assistance, the junior radiologist correctly diagnosed it as CL. In the CLP case, both AIOC and the senior radiologist correctly diagnosed it as CLP, while the junior misdiagnosed it. With AIOC assistance, the junior radiologist correctly diagnosed it as CLP. These cases highlight AIOC’s ability to support less experienced radiologists in making accurate diagnoses. Additionally, the AIOC system can further accelerate the diagnosis procedure of junior radiologists, reducing their diagnostic time to 5.31 seconds (6.62 seconds faster).

We further analyzed the automation bias of the use of AIOC for junior radiologists. We examined the OC-3000 across the three junior radiologists (R4, R5, and R6), comparing their diagnostic
performance when the AIOC provided correct versus incorrect recommendations. We measured overreliance as the percentage of cases where radiologists followed AI recommendations when the AI was incorrect, and appropriate reliance as the percentage when the AI was correct. As shown in Table~\ref{tab:automation_bias}, the results demonstrate a minor impact of automation bias on clinical decision-making. Overreliance rates were relatively low across all radiologists (mean: 9.8$\%$, range: 5.9-11.8$\%$), indicating that radiologists maintained critical judgment even when presented with incorrect AI suggestions.

\begin{figure}[!t]
    \centering
    \includegraphics[width=1\textwidth,height=\textheight,keepaspectratio]{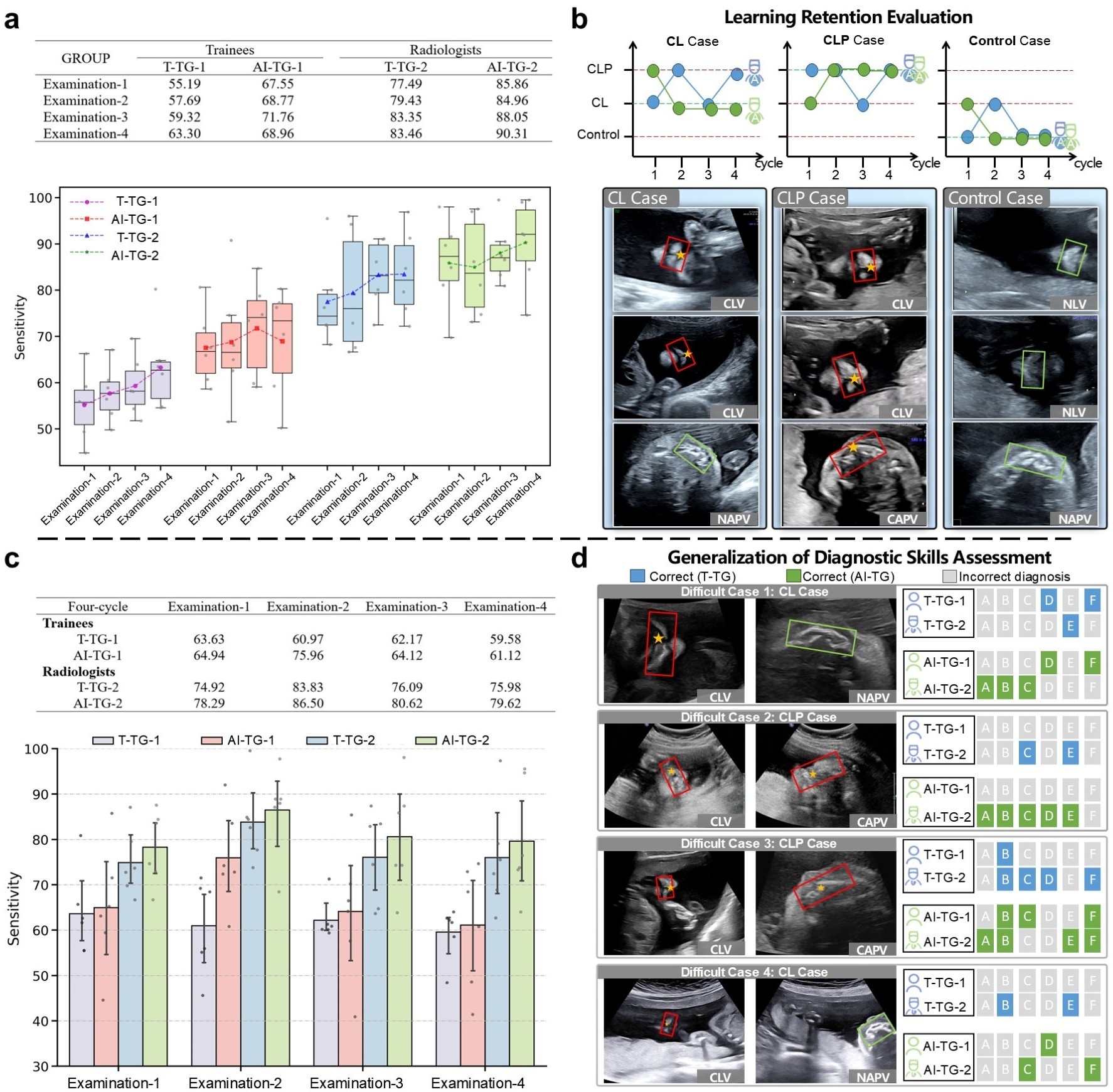}
    \caption{\textbf{Medical training results of the pilot trial involving radiologists and trainees in fetal OC and healthy case-based diagnosis.} (a) Detailed quantitative analysis of average sensitivity (\%) and box plots of trainees and radiologists in traditional training group (T-TG) and AI-augmented training group (AI-TG) during the four-cycle examination of 200 fixed cases. Box plots show the sensitivity of individual participants (n = 6 participants per box, represented by gray dots within each box). Box plots display median (center line), 25th–75th percentiles (box bounds), whiskers extending to the minimum and maximum values within 1.5 × IQR from the quartiles. (b) Learning curve of three specific cases (control, CL, and CLP) from the 200 fixed cases, diagnosed by one radiologist (T-TG-2-A, blue) from the T-TG group and one radiologist (AI-TG-2-A, green) from the AI-TG group. 
    (c) Detailed quantitative analysis of average sensitivity (\%) and bar plots of trainees and radiologists in two groups (T-TG and AI-TG) during the four-cycle examination of 100 random cases. Data are presented as the mean (measure of center) and 95\% confidence intervals represented by error bars. (d) Results from four challenging cases across four examinations. The left panels show representative ultrasound views of each case, with anatomical landmarks annotated. Green boxes indicate normal structures, red boxes and yellow stars mark abnormal structures and cleft locations, respectively. The right panels display diagnostic outcomes for six participants (A–F) in each group: blue indicates the correct diagnosis for the T-TG group, green for the AI-TG group, and grey indicates an incorrect diagnosis.
    CL: cleft lip; CLP: cleft lip and palate; CLV: cleft lip view; NLV: normal lip view; NAPV: normal alveolus and palate view; CAPV: cleft alveolus and palate view; T-TG: traditional training groups; AI-TG: AI training groups.
    }
    \label{fig: training}
\end{figure}

\subsection{Medical training pilot study}
In the medical education pilot of AIOC, we investigate the improvement of diagnostic capacity of multi-level participants (junior radiologists with 1-3 years of experience and trainees with 0-1 years of experience) with the help of the AIOC system. In the four training-and-exam cycle, we observed a consistent performance advantage in the AI-TG compared to the T-TG (Fig.~\ref{fig: training} (a) and (c)).

For the 200 fixed cases to assess learning retention, as shown in Extended Data Table~\ref{tab: S7-200}, AI-TG-1 demonstrated higher sensitivities (67.55-71.76$\%$) than T-TG-1 (55.19-63.30$\%$) across all examination cycles, with comparable specificities (89.98-90.56$\%$ vs 84.15-88.77$\%$) and improved accuracies (88.17-90.00$\%$ vs. 83.22-88.95$\%$). 
More pronounced improvements were observed in the radiologist groups. AI-TG-2 achieved superior diagnostic performance compared to T-TG-2 counterparts, with sensitivities of 85.86-90.31$\%$ vs 77.49-83.46$\%$, specificities of 95.00-97.79$\%$ vs 93.86-96.92$\%$, and accuracies of 94.11-97.78$\%$ vs 92.11-96.28$\%$ (Fig.~\ref{fig: training} (a) and Extended Data Table~\ref{tab: S7-200}).
This difference reached statistical significance on the third examination ($p$ = 0.03, $\alpha$ = 0.05). 
Besides, as shown in Fig~\ref{fig: training} (b), we track the diagnostic results of two junior radiologists with and without AI support, denoted as AI-TG-2-A and T-TG-2-A, on three cases with different types (i.e., CL, CLP, and Control) in the four-cycle examinations. It can be observed that with the assistance of AIOC, junior radiologists exhibited better learning retention. They were able to correct their diagnostic choices in the second round examination and maintain consistency in the follow-up rounds (green), while the one without AIOC support shows fluctuating clinical decisions (blue). According to the experimental record, AI-TG-2-A showed a clear improvement trend in accuracy across examination cycles (87.50$\%$, 95.50$\%$, 95.00$\%$, 96.00$\%$), while T-TG-2-A exhibited fluctuating performance (88.50$\%$, 87.00$\%$, 83.50$\%$, 94.00$\%$).

For the 100 random cases to evaluate generalization of diagnostic skills, the advantage of AI-TG was further validated and is shown in Fig.~\ref{fig: training} (c). AI-TG-1 achieved a higher average accuracy than T-TG-1 in the four-cycle examination (90.92$\%$ vs. 89.50$\%$, $p$ = 0.23), though the difference was not statistically significant. Additionally, AI-TG-2 demonstrated significantly higher accuracy than T-TG-2 (97.12$\%$ vs. 95.81$\%$, $p$ = 0.01, see Extended Data Table~\ref{tab: S7-200}). 
Furthermore, in Fig.~\ref{fig: training} (d), we selected four representative challenging cases from 100 random cases (one per cycle) and visualized the exam results of the four groups. The results indicate that the AI-TG generally showed better performance than T-TG on these challenging cases (details refer to Supplementary Table~\ref{tab:S8}).

\section{Discussion}\label{sec3}

Accurate prenatal ultrasound diagnosis of fetal OC is critical for preventing craniofacial disabilities and reducing infant mortality. In this study, we developed AIOC, an AI-powered diagnostic system that integrates deep learning with expert radiological knowledge to mimic the diagnostic workflows used by experienced clinicians. Our AIOC system demonstrated diagnostic performance comparable to that of senior radiologists across most metrics while significantly reducing interpretation time (Tables~\ref{tab:diagnostic_performance-oc3000} and~S5). Moreover, when used as a computer-aided diagnostic tool, AIOC substantially improved the clinical performance of junior radiologists, enhancing their sensitivity by 6.18$\%$. The consistent high accuracy of the AIOC system across both our internal OC-6000 dataset and external OC-3000 dataset (Table~\ref{tab:diagnostic_performance-oc3000}) demonstrates AIOC's superior generalizability across varied clinical scenarios derived from more than 20 hospitals. These results suggest the potential of AIOC as a reliable tool for high-quality diagnostic support that could be integrated into routine clinical workflows.

Early diagnosis of fetal OC enables crucial time-sensitive actions: intervention planning, genetic testing, identification of associated conditions, confirmation of overall fetal health, and psychological support for families~\cite{tonni2023cleft,wu2021prospective}. However, accurate detection before 20 weeks of gestation presents significant challenges, particularly for less experienced radiologists, due to subtle anatomical features and imaging limitations. Our study demonstrates that AIOC can effectively address this challenge, showing robust performance even in early gestational ages (14-17 weeks). Despite lacking specific training data for this period, the system achieved superior results on the OC-Early dataset (Table~\ref{tab:diagnostic_performance-oc3000}). This performance suggests that AI-augmented diagnosis could improve early detection capabilities, particularly for junior radiologists who traditionally struggle with these challenging cases.

Beyond early detection, our findings reveal the stability of the AIOC system across the full gestational spectrum (Fig.~\ref{fig: diagnosis_oc3000}(c)). This stability addresses another fundamental challenge in clinical practice, the rapid anatomical changes during mid-pregnancy that often compromise diagnostic consistency. This results in the variability of the diagnostic performance of the junior radiologists. The consistent performance of the presented AIOC system across different gestational ages suggests its potential to standardize diagnostic reliability throughout pregnancy, potentially reducing the documented variability in detection rates. This aspect of stability becomes particularly valuable when considering the need for consistent diagnostic capabilities across different clinical settings, where radiologists experience levels may vary significantly.

AI tools have advanced significantly, but their adoption in clinical practice, particularly for fetal diagnosis, remains limited due to lack of clinically meaningful outputs and suboptimal user experience~\cite{tang2023multicenter,zhang2022development,taksoe2024role,esteva2019guide}. Standard prenatal ultrasound examination for OC follows a systematic diagnostic protocol: radiologists first acquire standard fetal facial views, then systematically examine key anatomical structures (lips, alveolus, and palate), and finally integrate findings across multiple views to reach a case-level diagnosis. Unlike current AI models that provide isolated frame-by-frame detection results or standalone classification probabilities, our proposed AIOC system produces structured diagnostic outputs that align with this sequential decision-making process. This integration across multiple views is clinically essential. For example, confirming the CL diagnosis requires identifying a cleft lip in the coronal view and verifying normal alveolus and palate in the axial view. Our structured approach addresses the complexities of fetal anatomy analysis, where multiple anatomical confirmations are clinically relevant.

Our AIOC system demonstrated superior sensitivity in OC diagnosis across gestational ages of 14-28 weeks, albeit with slightly lower specificity (Table~\ref{tab:diagnostic_performance-oc3000}). 
Junior radiologists exhibited higher specificity than both senior radiologists and AIOC, likely due to their conservative diagnostic approach, which is caused by inexperience with detecting anomalies. In clinical practice, radiologists potentially intend to diagnose a fetus as healthy when confronted with cases showing subtle abnormalities, given the prior knowledge of low OC incidence in the general population. AIOC provided assistance by identifying case-specific issues and focusing on abnormal views and structures, thereby improving detection rates for junior radiologists.
Although the false positive rate is 3-5 times higher when AIOC is used alone, the AIOC is designed for assisted usage as a copilot, not standalone use. AI-assisted configuration can achieve optimal performance with both the lowest false positive rate and false negative rate compared to AIOC alone, as well as junior and senior radiologists.
Typically, balancing false negatives (leading to missed diagnoses and delayed intervention) and false positives (causing increased resource use and patient anxiety) is the key component when adapting the AI model to clinical practice~\cite{xue2020challenges}. Therefore, we recommend using AIOC as a copilot for clinical implementation, incorporating expert consultation for secondary confirmation. This approach balances improved detection sensitivity with appropriate specificity, minimizing unnecessary resource utilization and patient anxiety.

Most current studies on AI-assisted diagnosis typically use a clean dataset after strict manual quality control. However, this may limit the practicality of these models when applied to real-world clinical practice, because image quality and view orientation are often suboptimal. For instance, oblique views shown in Fig.~\ref{fig: diagnosis_oc6000}(d) occur occasionally due to the variability of radiologists' experience. The image selection in our study focused on whether the view image was sufficient to support OC diagnosis rather than on optimal image quality and orientation. Including those clinically imperfect images in AIOC training can enhance its generalizability. Critically, incorporating these images into the evaluation can further reflect the model's performance in real-world clinical practice.

Automation bias represents a significant concern in AI-assisted medical diagnosis, where over-reliance on algorithmic recommendations can potentially compromise clinical decision-making. While our findings suggest that radiologists maintained appropriate critical judgment when using AIOC, with overreliance rates remaining below 12$\%$ (Table~\ref{tab:automation_bias}), the broader implications of human-AI interaction in clinical settings warrant ongoing attention. Future clinical implementation should incorporate appropriate training protocols and monitoring mechanisms to ensure AI systems serve as diagnostic aids rather than decision replacements.

Our findings also reveal the educational potential of the AIOC system in developing clinical expertise for OC diagnosis. In our four-cycle training pilot study, AI-augmented training groups consistently outperformed traditional training groups in diagnostic performance. Radiologists with access to AIOC-supported training demonstrated improved diagnostic accuracy and more consistent performance improvement across examination cycles (Extended Data Table~\ref{tab: S7-200}, Fig.~\ref{fig: training}). Combining view classification, structural identification, and diagnostic reasoning, the AIOC system provides a systematic approach to the expertise development in rare condition diagnosis. This dual functionality of AIOC, a joint diagnostic aid and educational framework, addresses the challenge in medical training for rare conditions, where limited case exposure traditionally restricts skill development~\cite{cash2001accuracy,paradowska2022current,mossey2023global}. 
AIOC provides the potential to transform each clinical encounter into a structured training component, which can be particularly valuable in resource-limited settings, where specialized training opportunities are scarce~\cite{baeza2024diagnostic,taksoe2021simulation,mullaney2019qualitative,lagrone2012review} but could be supplemented with AI-augmented learning tools to help close expertise gaps between junior and senior radiologists~\cite{agbenorku2013orofacial,lei2024development,rosman2019developing}.

The implications of the AIOC system demonstrate a potential solution for rare disease diagnosis globally~\cite{esteva2019guide}. The AIOC system bridges the expertise gaps in resource-limited settings by embedding expert-level diagnostic capabilities into accessible AI tools. This approach is particularly relevant for conditions like OC, where early intervention significantly improves outcomes, while detection rates vary based on regional healthcare resources~\cite{beaty2016genetic, world2022global,hannun2019cardiologist}. Furthermore, the educational framework embedded within AIOC could accelerate expertise development in underserved regions, potentially expanding diagnostic capacity beyond traditional training limitations. This combination of immediate diagnostic assistance and long-term expertise development offers a promising pathway toward more equitable healthcare delivery for rare conditions globally.

While Grad-CAM visualization demonstrates that our model captures anatomically meaningful features in successful cases (Fig.~\ref{fig: diagnosis_oc6000} (d)~and~\ref{fig: diagnosis_early} (b)), analysis of failure modes provides additional insights into the model's reasoning process. Examination of misclassified cases (Supplementary Fig.~\ref{fig: correct_incorrect}) illustrates that attention patterns in failed cases remain focused on anatomically relevant regions, but with subtle yet critical differences. For example, in cases where CLP was misclassified as CL, the model correctly identified the cleft lip but mislocalized attention to the intact portion of the palate rather than the cleft area. This fine-grained attention error demonstrates that while our model generally focuses on appropriate anatomical structures, distinguishing subtle variations within these regions remains challenging and represents a key area for future improvement.

Future works can be intended from the following aspects. First, although our study recruited a large-scale multi-center dataset, the diversity remains limited in several aspects, including a small number of cases in the first trimester (with only one CL fetus in the early-timester subset), single-ethnicity representation, and limited OC subtypes. Furthermore, our inclusion criterion requiring complete key views may introduce selection bias toward higher-quality exams, which may not reflect the full spectrum of real-world screening conditions. These limitations may affect the model's generalizability when applied to other populations or clinical settings in real practice. 
Additionally, although our collected dataset spans 22 hospitals across the OC-6000 and OC-GT3000 subsets, the external validation set is not fully hospital-disjoint from the internal set, and data splits were not fully stratified by hospital or device, which may result in some degree of site-style leakage. The ultrasound equipment distribution of the dataset is also skewed toward dominant platforms, and performance on under-represented devices may remain less certain.
Future work will focus on expanding dataset diversity by incorporating a broader range of gestational ages (including the first trimester), covering additional OC types (e.g., isolated cleft palate), and including multi-ethnic populations. Secondly, although our model achieved good performance on the collected retrospective dataset, its true performance in real-world clinical settings remains to be validated. We plan to conduct prospective studies to evaluate the AIOC's performance in clinical practice, with a particular focus on early diagnosis in the first trimester, comprehensive diagnosis of OC subtypes, analysis of dynamic video data, and the model's generalizability across different settings, devices, and populations.
Additionally, emphasizing the interpretability of AI systems by providing clearer insights into decision-making processes and reducing automation bias is crucial for building trust and promoting clinical adoption. These advancements aim to empower clinicians to rely more confidently on AI assistance in daily practice while improving their diagnostic skills. Moving forward, we will continuously update our AIOC system to establish it as a true clinical assistive tool, capable of providing early and accurate OC diagnosis throughout the entire pregnancy. This will help improve diagnostic outcomes and address challenges such as junior radiologists' limited skills, data scarcity, and fetal complexity, particularly in low- and middle-income countries~\cite{world2022global}.

\section{Methods}\label{sec2}

\subsection{Ethical approval}

This study was approved by the Clinical Research Ethics Committee of the Affiliated Suzhou Hospital of Nanjing Medical University (K-2023-067-H01) and conducted following the Declaration of Helsinki. It was registered at www.chictr.org.cn (ChiCTR2300071300). Due to its retrospective design, written informed consent was waived. All patient data has been fully anonymized to prevent any potential re-identification.

\subsection{Data collection}

This multicenter, retrospective study collected data from 9,215 fetuses, comprising 45,139 ultrasound images of both OC and healthy cases between gestational weeks 14–28 (mean$\pm$std: 22.5$\pm$2.1). The study was conducted across 22 hospitals coordinated by the Affiliated Suzhou Hospital of Nanjing Medical University from January 2018 to December 2023 (Extended Data Table \ref{tab: S1}).
The key inclusion criteria comprised three aspects. 
(a) Only women with a confirmed singleton pregnancy were included in the study.
(b) Each case was required to contain the necessary ultrasound views for clinical diagnosis. 
(c) Postnatal follow-up data were available for fetuses.

\subsection{Ultrasound Image Acquisition and Quality Control}

All fetal examinations were performed by radiologists certified in prenatal diagnosis. Over 100 experienced radiologists, each with 8–30 years of expertise in fetal US screening and diagnosis, conducted real-time US scanning, captured key diagnostic images, and provided diagnoses for each case. When abnormalities were suspected, a senior radiologist with over 10 years of experience in fetal US diagnosis was consulted for diagnostic confirmation. The final diagnosis for each fetus was verified against postnatal outcomes as the gold standard. All pregnant women were of Chinese ethnicity with an average age of 29.3$\pm$6.8 years.

All ultrasound examinations were performed using high-end equipment (GE Voluson E8/E10, GE HealthCare; Philips EPIQ7/Affiniti 70, Philips Healthcare; Samsung UGEO WS80A, Samsung Medison, See Table~\ref{tab: S1} for details). Image acquisition followed the standards established by the International Society of Ultrasound in Obstetrics and Gynecology and the Chinese Medical Association expert consensus. For each fetus, facial views were obtained in both axial planes (demonstrating alveolar and palatal anatomy) and coronal planes (showing nasal, upper lip, and lower lip structures).

To facilitate precise classification, axial planes were categorized as NAPV and  CAPV, while coronal planes were classified as NLV and CLV. The diagnostic key views were NAPV and NLV for control cases, NAPV and CLV for CL cases, and CAPV and CLV for CLP cases.
Quality control was rigorously conducted by three radiologists with over 20 years of fetal ultrasound experience to ensure the presence of adequate diagnostic key views for each case. As this was a retrospective study, strict selection based on image gain, magnification, or display orientation was not enforced to better reflect real-world clinical conditions.

\subsection{Dataset Construction and Partitioning}
To develop and validate the AI-assisted OC (AIOC) system, we constructed three datasets: OC-6000, OC-GT3000, and OC-Early.
The internal dataset, OC-6000, was collected from 19 hospitals and included 28,994 images from 6,010 fetuses (gestational age 18–28 weeks), comprising 5,078 control fetuses (22,855 images), 103 CL fetuses (678 images), and 829 CLP fetuses (5,461 images). The dataset was partitioned at the case level into training, validation, and testing sets following a ratio of 80$\%$, 10$\%$, and 10$\%$, yielding 4,168, 415, and 495 control fetuses; 64, 18, and 21 CL fetuses; and 684, 75, and 70 CLP fetuses, respectively.

The external validation dataset, OC-GT3000, was collected from 12 hospitals and comprised 15,848 images from 3,168 fetuses (gestational age 18–28 weeks), including 2,980 control fetuses (14,674 images), 18 CL fetuses (128 images), and 170 CLP fetuses (1,046 images).
The early gestation dataset, OC-Early, was collected from 5 hospitals and included 297 images from 37 fetuses at 14–17 weeks of gestation, comprising 18 control fetuses (51 images), 1 CL fetus (13 images), and 18 CLP fetuses (233 images).

To mimic clinical practice in OC diagnosis, five radiologists with 5–10 years of experience annotated the OC-6000 dataset using the Pair annotation software package~\cite{liang2022sketch}. For each case, the radiologists annotated specific views and key structures based on the diagnosis (Control, CL, CLP). Control cases were annotated with NLV and NAPV views, marking bounding boxes around the upper lip and alveolar ridge; CL cases with CLV and NAPV views, marking bounding boxes around the cleft lip and normal alveolar ridge; and CLP cases with CLV and CAPV views, marking bounding boxes around the cleft lip, cleft alveolus, and cleft palate. Random checks and adjustments were performed by three experienced radiologists to ensure annotation accuracy. Finally, another radiologist (Y.Z.), with 6 years of experience and no involvement in the annotation process, organized the annotated data for model training.

\subsection{Model Development}
Our model integrates detection and classification branches to enhance diagnostic accuracy, efficiency, and interpretability. 
During training, the YOLOX-based branch~\cite{zheng2021yolox} was adopted for its improved detection accuracy, inference speed, and generalization capabilities.
Specifically, it takes the image as input and outputs the rotated boxes of structures. 
This process is driven by the classification, bounding box, and objectness losses. Specifically, the classification loss is to measure the difference between the predicted probabilities ($\hat{p}_i$) and GT categories ($y_i$):
\begin{equation}
    L_{\mathrm{cls}} = - \sum_{i = 1}^{N} y_i \log (\hat{p}_i) ,
\end{equation}
where $N$ is the number of categories. 
Besides, the Intersection over Union (IoU) based loss is adopted to optimize the predicted boxes. Here, the Generalized IoU (GIoU)~\cite{rezatofighi2019generalized} was selected to improve the traditional IoU calculation (e.g., non-overlap penalization, avoidance of gradient vanishing, etc.), with the formula as:
\begin{align}
\mathrm{IoU} &= \frac{\left|A \cap B\right|}{\left|A \cup B\right|}, \\
\mathrm{GIoU} &= \mathrm{IoU} - \frac{\left|C - (A \cup B)\right|}{\left|C\right|}, \\
L_{\mathrm{IoU}} &= 1 - \mathrm{GIoU} ,
\end{align}
where $A$ and $B$ are two rotated boxes, $A \cap B$ and $A \cup B$ represents their intersection and union areas, respectively. $C$ denotes the area of the minimum external box of $A$ and $B$. We also introduced the binary cross-entropy loss to judge whether the predicted boxes contain targets or not, which can be written as:
\begin{equation}
L_{\mathrm{bce}} = - \sum_{i = 1}^{N} y_i \log (\hat{p}_i)+(1 - y_i)\log (1 - \hat{p}_i).
\end{equation}

Additionally, an L2-based loss function, named $L_{\mathrm{ratio}}$, was employed to calculate the area ratio between the rotated and horizontal boxes, effectively enhancing geometric learning within our proposed AIOC system. 
In summary, the total loss function for the detection branch can be defined as: 
\begin{equation}
L
=
L_{\mathrm{cls}}
+
L_{\mathrm{IoU}}
+
L_{\mathrm{bce}}
+
L_{\mathrm{ratio}}.
\end{equation}

For the classification branch, we used the Mamba-Inspired Linear Attention (MILA) framework for its validated satisfactory performance in different classification tasks~\cite{han2024demystify}. 
It combined linear attention and Mamba designs to reduce computation complexity while enhancing model representation ability. 
Specifically, it first extracted global features from input images. Then, according to the detection results, the local features corresponding to the key structures are cropped from the feature map. These features are resized to concatenate with the global one, obtaining the features of size ($k$+1)*512, where $k$=5 is the number of local features. When the number of detected key structures is less than 5, blank features will be added. We further equip a Long Short-Term Memory (LSTM) module to model the interrelationships among structural features and output the classification results. The training procedure is optimized using the conventional cross-entropy loss. Ultimately, a threshold-based post-processing strategy will be integrated with expert knowledge, fusing image-based results to conduct case-based diagnosis.

Experiments were conducted on one NVIDIA GeForce RTX 4090 GPU, utilizing PyTorch (version 2.0.0). The workflow involved resizing images to $480\times480$ pixels for detection and $224\times224$ pixels for classification. The detection branch effectively identified five key structures, with parameters $\{x_1, y_1, x_2, y_2, w, h, \theta\}$. $\{x_1, y_1\}$ and $\{x_2, y_2\}$ are the left-top and right-down point coordinates. w and h are the offsets from the horizontal box to the rotated one; $\theta$ is the area ratio. The SGD optimizer was used, with cosine annealing strategy to adjust the learning rate (1$e^{-3}$ for initial). For the classification branch, it was optimized by AdamW and categorized images into four view types. The learning strategy adopted is similar to that used in detection. For both tasks, training epochs are 100. Diagnostic outputs were derived from a joint analysis of views and structures per case, facilitating the diagnosis of control, CL, and CLP cases. The development process of the AIOC system is detailed in Fig. \ref{fig: overview}(a).

\subsection{Clinical Evaluation Studies and Medical Training Pilot}

For both the clinical evaluation study and the medical training pilot, data were managed using the EDS platform with a custom web-based interface for double-blind case presentation and diagnosis. All patient information was anonymized. Participants were blinded to group allocation during clinical evaluation. Each participant was assigned a unique coded identifier and prohibited from communicating with others. Examinations were conducted in fixed hospital diagnostic rooms at consistent weekly intervals with standardized case numbers per participant.

For the clinical evaluation study, we recruited six radiologists: three senior radiologists (\textgreater 10 years of experience, R1-R3) and three junior radiologists (3 years of experience, R4-R6). We evaluated the independent diagnostic capabilities of both radiologists and AIOC in distinguishing OCs from healthy fetuses on the OC-GT3000 dataset. The study was conducted over three weeks with approximately 1,000 cases per session (Control:CLP:CL = 888:120:22). Participants were given 24 hours to complete each session. All diagnoses were validated against postnatal ground truth. In the AI-assisted diagnosis evaluation phase, junior radiologists were provided with AIOC-generated visualizations of anatomical planes, highlighted structures, and preliminary diagnostic results. They then make diagnoses by considering the AI-provided information, enabling evaluation of performance improvements with AI assistance. A washout period of two weeks was implemented between the unassisted and AI-assisted diagnosis phases for junior radiologists.

For the medical training pilot study, twenty-four physicians participated, including 12 trainees with 0-1 year of experience and 12 junior radiologists with 1-3 years of experience. All participants first completed a theoretical lecture on OCs delivered by a senior radiologist with over 15 years of experience, followed by a basic test using 20 cases from the OC-6000 dataset (Control:CLP:CL = 15:4:1). The lecture focused on the causes of various OC abnormalities and representative ultrasound images for each condition. Participants were then randomized into two groups stratified by experience level: the Traditional Training Group (T-TG) and AI Training Group (AI-TG). This resulted in four subgroups: T-TG-1 and AI-TG-1 (junior radiologists, n=6 each), and T-TG-2 and AI-TG-2 (trainees, n=6 each).

The training program consisted of four cycles, with a two-week washout period between each. Each cycle included two phases: (1) a training phase with 20 novel cases (Control:CLP:CL = 15:4:1), and (2) an examination phase with 300 cases from the OC-3000 dataset. During the training phase, T-TG participants reviewed original images with reference diagnoses, while AI-TG participants additionally reviewed AIOC-generated analyses, including view categories, bounding boxes of anatomical structure, and AI diagnostic recommendations. The examination phase comprised 200 fixed cases that remained consistent across all cycles (Control:CLP:CL = 125:69:6) to assess learning retention, and 100 random cases (Control:CLP:CL = 72:25:3) to evaluate diagnostic performance with new presentations. All participants independently diagnosed examination cases without AI support within a 3-hour time limit per session.

\subsection{Evaluation Metrics and Statistical Analysis}

The performance metrics for case-based diagnosis between the AIOC system and radiologists were evaluated using accuracy, sensitivity, specificity, FNR, FPR, F1-score, Youden index, and confusion matrix. 
Specifically, AUC is defined as the area under the ROC curve. 
The Youden Index is a balanced summary metric ranging from -1 to 1.
Given the imbalanced distribution of normal and abnormal cases in our dataset, it is employed to provide an unbiased measurement for the AIOC system and radiologists.
All metrics can be formally defined as:

\begin{equation}
\mathrm{Accuracy}
=
\frac{\mathrm{TP} + \mathrm{TN}}
{\mathrm{TP} + \mathrm{TN} + \mathrm{FP} + \mathrm{FN}},
\label{eq:acc}
\end{equation}

\begin{equation}
\mathrm{Sensitivity}
=
\frac{\mathrm{TP}}
{\mathrm{TP} + \mathrm{FN}},
\label{eq:sen}
\end{equation}

\begin{equation}
\mathrm{Specificity}
=
\frac{\mathrm{TN}}
{\mathrm{TN} + \mathrm{FP}},
\label{eq:spe}
\end{equation}

\begin{equation}
\mathrm{FNR}
=
\frac{\mathrm{FN}}
{\mathrm{FN} + \mathrm{TP}},
\label{eq:fnr}
\end{equation}

\begin{equation}
\mathrm{FPR}
=
\frac{\mathrm{FP}}
{\mathrm{FP} + \mathrm{TN}},
\label{eq:fpr}
\end{equation}

\begin{equation}
\mathrm{F1\text{-}score}
=
\frac{2\,\mathrm{TP}}
{2\,\mathrm{TP} + \mathrm{FP} + \mathrm{FN}},
\label{eq:f1}
\end{equation}

\begin{equation}
\mathrm{Youden\ Index}
=
\mathrm{Sensitivity}
+
\mathrm{Specificity}
-
1,
\label{eq:J}
\end{equation}
where $\mathrm{TP}$, $\mathrm{FP}$, $\mathrm{TN}$, and $\mathrm{FN}$ denote the number of
true positives, false positives, true negatives, and false negatives in the confusion matrix, respectively.

Additionally, a chi-square test was performed to assess statistically significant differences in performance between the AIOC system and radiologists. SD was used to evaluate the stability of the AIOC system across gestational ages. A \textit{t}-test was conducted to determine statistically significant differences in time cost between the AIOC system and the radiologists. All analyses were performed using Python (version 3.11.0), with a \textit{p}-value \textless 0.05 considered statistically significant.

\section{Data Availability}
Due to privacy restrictions, patient data is protected and can only be accessed with the consent of the data management committee at institutions; it is not publicly available. Requests for non-commercial use of the data and relevant clinical information should be sent to the corresponding authors. The request will be reviewed by the data management committee and informed to the applicant within one month. 

\section{Code Availability}
The code for developing the algorithm is provided via Zenodo at \url{https://doi.org/10.5281/zenodo.18805366}~\cite{huang_2026_18805366}.

\bibliography{export}

\newpage
\section*{Acknowledgements}
This work was supported by the Science and Technology Planning Project of Guangdong Province (No. 2023A0505020002, D.N.), the National Natural Science Foundation of China (Nos. 12326619, 62171290, D.N.; No. 62572324, Yuhan.Z.), the Frontier Technology Development Program of Jiangsu Province (No. BF2024078, D.N.), the Suzhou General Project for Science, Education and Health Strengthening (No. MSXM2024022, L.Y.), the Guangdong Province Key Areas Research and Development Program Project (No. 2020B1111130002, Y.X.), the Clinical Research Project of Shenzhen Luohu People's Hospital (No. LCYJ202202, Y.X.), the Guangxi Key Research and Development Program (No. AB23026042, H.Z.), the Multi-center clinical study of intelligent prenatal ultrasound (No. ChiCTR2300071300, X.Z.), the Royal Academy of Engineering, United Kingdom under the RAEng Chair in Emerging Technologies (INSILEXCiET1919/19, A.F.F.), the ERC Advanced Grant UKRI Frontier Research Guarantee (INSILICO EP/Y030494/1, A.F.F.), the UK Centre of Excellence on in-silico Regulatory Science and Innovation (UK CEiRSI) (10139527, A.F.F.), the National Institute for Health and Care Research (NIHR) Manchester Biomedical Research Centre (BRC) (NIHR203308, A.F.F.), the BHF Manchester Centre of Research Excellence (RE/24/130017, A.F.F.), and the CRUK RadNet Manchester (C1994/A28701, A.F.F.).

The authors gratefully acknowledge the involvement of 22 hospitals in this multicenter study. Special thanks are extended to the clinicians from the participating institutions for their clinical support, including Haining Chen, Jiajia Qu, and Yueyue Xu from The People’s Hospital of Guangxi Zhuang Autonomous Region; Yan Cheng from Hangzhou Women’s Hospital; Li Lv from Northwest Women’s and Children’s Hospital; Chunyun Zhang, Chunli Lu, Fangyi Deng, Dan Xie, Li Yang, Wenxue Pei, Zhiying Xie, and Meilun Pan from Shenzhen Guangming District People’s Hospital; Jiaxuan Yang, Minjin Lin, Shuping Lin, Juan Liu, Guanliang Li, and Erbao Li from Shenzhen Luohu People’s Hospital; and Shumian Xu, Naimin Sun, Chengcheng Wu, Yingqi Fang, Jihong Peng, and Yue Zhao from The Affiliated Suzhou Hospital of Nanjing Medical University. Additional acknowledgment goes to the data processing and model development team: Gaocheng Cai, Kaiying Wang, and Ran Li from Shenzhen RayShape Medical Technology Co., Ltd.; Shuai Li, Yanlin Chen, Huanwen Liang, Xing Tao, Jian Wang, Jingxian Xu, Zhaojin Chen, Yang Yang, and Hongzhang Wang from the Medical Ultrasound Image Computing (MUSIC) Laboratory, Shenzhen University.

\section*{Author Contributions}

Yuanji Z., Y.H., and H.D. participated in conceptualization, methodology, data analysis, and writing of the original draft. X.Z. participated in methodology, data analysis, and writing of the original draft.
G.H., X.Y., Y.X., L.Y., X.D., and D.N. participated in conceptualization, supervision, and funding acquisition.
C.L., Z.Y., L.L., Jiuping L., H.Z., X.G., C.Y., L.S.,  participated in conducting the clinical reader study.
S.L., R.L., Y.C., Yuhan Z., Jiewei L., Yongsong Z., M.Y., H.L., X.H., C.C., J.Z., W.P., and A.F.F provided critical comments and reviewed the paper. All authors contributed to the research, editing, and approval of the paper.

\section*{Competing Interests}
The authors declare that they have no known competing financial interests or personal relationships that could have appeared to influence the work reported in this paper.

\newpage
\section*{Extended Data}

\setcounter{table}{0}  
\setcounter{figure}{0}
\setcounter{section}{0}
\setcounter{equation}{0}
\renewcommand{\thetable}{E\arabic{table}}
\renewcommand{\thefigure}{E\arabic{figure}}
\renewcommand{\thesection}{E\arabic{section}}
\renewcommand{\theequation}{E\arabic{equation}}

\begin{figure}[!h]
    \centering
    \includegraphics[width=1\textwidth,height=\textheight,keepaspectratio]{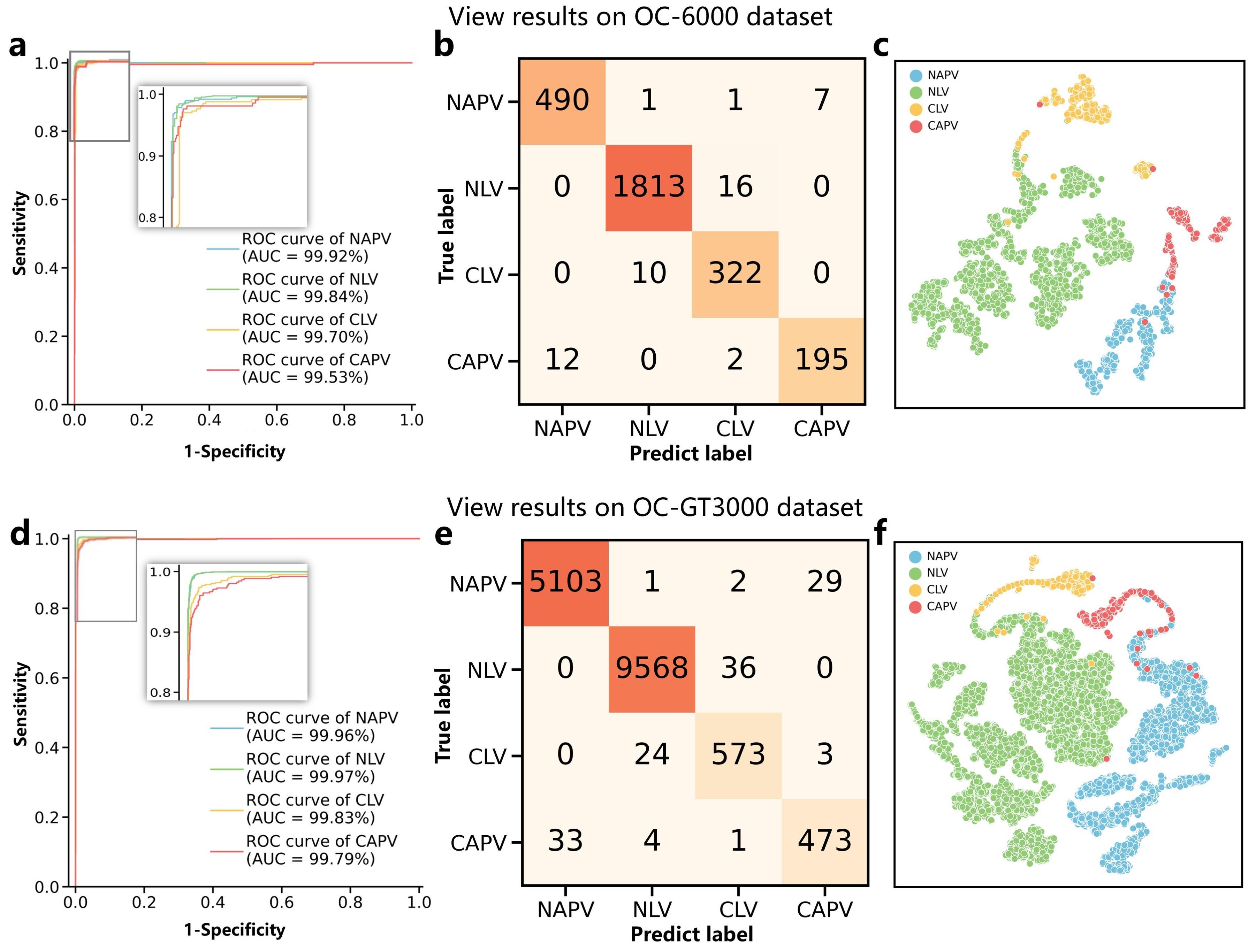}
    \caption{\textbf{View classification visualization of the AIOC system on the OC-6000 and OC-GT3000 datasets.} a and d: Confusion matrix of the view classification of the OC-6000 and OC-GT3000 datasets; b and e: Receiver operating characteristic (ROC) curve of the OC-6000 and OC-GT3000 datasets; c and f: t-distributed Stochastic Neighbor Embedding (t-SNE) of the OC-6000 and OC-GT3000 datasets. CLV: cleft lip view; NLV: normal lip view; NAPV: normal alveolus and palate view; CAPV: cleft alveolus and palate view; AUC: area under the curve; ROC: receiver operating characteristic.}
    \label{fig: S1}
\end{figure}

\begin{table}[h]
\centering
\caption{\textbf{Diagnostic performance of senior radiologists, junior radiologists, and AIOC-assisted junior radiologists across the three types of diagnoses: control, cleft lip (CL), and cleft lip and palate (CLP), in the OC-GT3000 dataset.} 
AUC: area under the receiver operating characteristic curve; CI: confidence interval; FPR: false positive rate; FNR: false negative rate; F1: F1-score. }
\label{tab: S5}
\begin{tabular}{lcccccccc}
\toprule
\textbf{Characteristic} & \textbf{AUC (\%)} & \textbf{Sensitivity (\%)} & \textbf{Specificity (\%)} & \textbf{Accuracy} & \textbf{FNR} & \textbf{FPR} & \textbf{F1} & \textbf{Youden} \\
 & \textbf{(95\%CI)} & \textbf{(95\%CI)} & \textbf{(95\%CI)} & \textbf{(\%)} & \textbf{(\%)} & \textbf{(\%)} & \textbf{(\%)} & \textbf{Index} \\
\midrule
\textbf{Senior R1} & & & & & & & & \\
\quad Control & 98.7(97.75-99.73) & 99.87(97.31-100.00) & 97.87(95.81-99.94) & 99.87 & 0.13 & 2.13 & 99.87 & 0.98 \\
\quad CL & 94.37(86.62-99.94) & 86.89(74.37-100.00) & 99.84(99.79-100.00) & 76.19 & 11.11 & 0.16 & 82.05 & 0.89 \\
\quad CLP & 98.19(90.92-99.95) & 99.89(97.31-100.00) & 99.9(99.79-100.00) & 98.2 & 3.53 & 0.1 & 77.33 & 0.96 \\
\quad Average & 97.11(94.37-99.69) & 95.50(89.27-100.00) & 99.2(98.43-99.97) & 91.42 & 4.92 & 0.8 & 93.08 & 0.94 \\
\hline
\textbf{Senior R2} & & & & & & & & \\
\quad Control & 98.9(97.90-99.71) & 99.87(99.51-100.00) & 97.87(95.81-99.94) & 99.87 & 0.1 & 2.13 & 99.88 & 0.98 \\
\quad CL & 88.81(77.33-93.96) & 77.78(53.87-99.98) & 99.84(99.79-100.00) & 75.68 & 22.22 & 0.16 & 75.68 & 0.78 \\
\quad CLP & 91.86(82.62-97.37) & 96.47(93.37-99.84) & 99.87(99.74-100.00) & 97.62 & 3.53 & 0.13 & 97.04 & 0.96 \\
\quad Average & 95.29(90.68-98.81) & 91.38(84.02-98.74) & 99.19(98.42-99.97) & 90.39 & 8.62 & 0.81 & 90.87 & 0.91 \\
\hline
\textbf{Senior R3} & & & & & & & & \\
\quad Control & 98.34(96.89-99.48) & 99.87(99.33-100.00) & 96.81(94.3-99.92) & 99.8 & 0.13 & 3.19 & 99.83 & 0.97 \\
\quad CL & 91.57(82.08-99.91) & 83.33(67.12-100.00) & 99.81(99.66-99.96) & 71.43 & 16.67 & 0.19 & 76.92 & 0.83 \\
\quad CLP & 97.91(96.73-99.33) & 95.03(89.29-98.87) & 99.93(99.94-100.00) & 98.79 & 4.12 & 0.07 & 97.31 & 0.96 \\
\quad Average & 95.4(91.74-99.57) & 93.08(82.93-96.63) & 99.8(93.90-99.76) & 90.01 & 6.97 & 1.15 & 91.36 & 0.92 \\
\hline
\textbf{Junior R4} & & & & & & & & \\
\quad Control & 99.2(98.21-100.00) & 100.0(100.0-100.00) & 98.4(96.1-100.00) & 99.9 & 0 & 1.6 & 99.95 & 0.98 \\
\quad CL & 88.67(76.97-97.29) & 77.78(53.87-96.98) & 99.75(97.59-99.92) & 63.64 & 22.22 & 0.25 & 70 & 0.78 \\
\quad CLP & 97.3(95.59-98.83) & 94.71(91.34-98.07) & 99.93(98.41-100.00) & 98.77 & 5.29 & 0.07 & 96.7 & 0.94 \\
\quad Average & 95.09(90.26-98.87) & 90.83(83.34-95.85) & 99.39(98.84-99.97) & 87.44 & 9.17 & 0.64 & 88.88 & 0.90 \\
\hline
\textbf{Junior R5} & & & & & & & & \\
\quad Control & 99.2(98.48-99.90) & 99.78(99.59-100.00) & 98.84(97.74-100.00) & 99.93 & 0.22 & 1.06 & 99.82 & 0.99 \\
\quad CL & 88.56(77.14-96.72) & 93.33(89.3-100.00) & 99.4(99.13-99.67) & 42.42 & 3.33 & 0.6 & 54.13 & 0.77 \\
\quad CLP & 97.03(95.98-98.72) & 94.12(90.58-97.65) & 99.93(98.41-100.00) & 98.77 & 5.88 & 0.07 & 96.39 & 0.94 \\
\quad Average & 94.98(90.41-98.81) & 95.74(91.23-99.28) & 99.42(98.81-99.89) & 80.37 & 3.14 & 0.58 & 83.7 & 0.90 \\
\hline
\textbf{Junior R6} & & & & & & & & \\
\quad Control & 99.57(99.36-99.75) & 99.43(99.16-99.77) & 99.02(98.15-99.83) & 98.99 & 0.57 & 7.98 & 99.46 & 0.91 \\
\quad CL & 69.21(58.05-81.31) & 38.89(19.16-61.41) & 99.52(98.15-99.78) & 99.18 & 61.11 & 0.48 & 35.0 & 0.38 \\
\quad CLP & 93.82(91.24-96.34) & 88.24(91.66-95.08) & 99.44(88.15-99.67) & 98.3 & 11.76 & 0.6 & 88.76 & 0.87 \\
\quad Average & 86.25(81.05-100.00) & 75.52(99.10-84.74) & 99.03(98.15-99.44) & 98.99 & 24.48 & 3.02 & 74.41 & 0.73 \\
\hline
\textbf{Junior-AIOC R4} & & & & & & & & \\
\quad Control & 99.9(99.71-100.00) & 99.93(99.43-100.00) & 99.47(98.43-100.00) & 99.97 & 0.07 & 0.53 & 99.95 & 0.99 \\
\quad CL & 94.38(85.64-99.97) & 88.89(74.37-100.00) & 99.87(99.75-100.00) & 80 & 11.11 & 0.13 & 84.21 & 0.89 \\
\quad CLP & 98.08(90.55-99.94) & 98.24(96.26-100.00) & 99.93(98.24-100.00) & 98.82 & 1.76 & 0.07 & 98.53 & 0.98 \\
\quad Average & 97.77(94.27-99.99) & 95.69(90.16-100.00) & 99.7(99.34-100.00) & 92.93 & 4.31 & 0.24 & 94.23 & 0.95 \\
\hline
\textbf{Junior-AIOC R5} & & & & & & & & \\
\quad Control & 99.17(98.44-99.87) & 99.87(99.44-100.00) & 98.64(96.1-100.00) & 99.9 & 0.13 & 1.06 & 99.91 & 0.98 \\
\quad CL & 91.15(81.77-99.00) & 93.33(63.12-100.00) & 99.84(99.85-100.00) & 88.24 & 16.67 & 0.03 & 85.71 & 0.83 \\
\quad CLP & 96.89(93.02-99.48) & 94.23(88.76-99.81) & 99.97(99.44-100.00) & 99.41 & 5.77 & 0.03 & 99.49 & 0.99 \\
\quad Average & 95.8(93.09-98.47) & 95.81(90.16-100.00) & 99.4(98.61-100.00) & 95.9 & 7.52 & 0.37 & 95.03 & 0.95 \\
\hline
\textbf{Junior-AIOC R6} & & & & & & & & \\
\quad Control & 99.63(98.07-99.93) & 99.66(99.46-99.87) & 99.78(96.61-100.00) & 99.59 & 0.34 & 1.6 & 99.78 & 0.98 \\
\quad CL & 86.0(74.87-95.84) & 72.22(49.46-92.91) & 98.4(96.1-100.00) & 99.62 & 27.78 & 0.22 & 68.42 & 0.72 \\
\quad CLP & 94.98(97.07-99.38) & 98.24(99.46-100.00) & 99.73(96.61-99.92) & 99.65 & 1.76 & 0.27 & 96.81 & 0.98 \\
\quad Average & 93.67(90.27-98.53) & 90.04(99.46-99.76) & 99.2(98.43-99.95) & 99.62 & 9.96 & 0.69 & 88.34 & 0.89 \\
\bottomrule
\end{tabular}
\end{table}

\begin{table}[!h]
\centering
\caption{The Diagnostic performance between Traditional training group (T-TG) and AI-augmented training group (AI-TG) in the pilot trial of 200 fixed cases and 100 random cases; The difference in sensitivity between T-TG and AI-TG was assessed using a two-sided Mann–Whitney U test. To control for family-wise error rate, $p$-values were adjusted for multiple comparisons using the Šidák correction. T-TG-1 and AI-TG-1: trainees; T-TG-2 and AI-TG-2: radiologists;  CI: confidence interval.}
\label{tab: S7-200}
\begin{tabular}{lccccccc}
\toprule
GROUP & Sensitivity & 95\%CI & Specificity & 95\%CI & Accuracy & 95\%CI & $p$-value\\
\midrule
 & & & \multicolumn{2}{c}{\textbf{200 fixed cases}}   & & & \\
\textbf{Examination-1} & & & & & & & \\
\quad T-TG-1 & 55.19 & 44.67 - 66.25 & 84.15 & 78.5 - 89.76 & 83.22 & 78.19 - 88.25 & \\
\quad AI-TG-1 & 67.55 & 64.41 - 90.99 & 89.98 & 90.23 - 97.36 & 88.17 & 88.51 - 95.71 & 0.17 \\
\quad T-TG-2 & 77.49 & 52.10 - 82.93 & 93.86 & 85.08 - 94.79 & 92.11 & 83.88 - 92.45 & \\
\quad AI-TG-2 & 85.86 & 74.46 - 95.66 & 95.00 & 91.82 - 98.00 & 94.11 & 91.0 - 97.14 & 0.60 \\

\textbf{Examination-2} & & & & & & & \\
\quad T-TG-1 & 57.69 & 47.92 - 69.33 & 86.73 & 81.21 - 92.21 & 86.28 & 81.65 - 90.91 & \\
\quad AI-TG-1 & 68.77 & 69.57 - 90.64 & 90.56 & 91.65 - 98.58 & 90.00 & 91.62 - 97.72 & 0.25 \\
\quad T-TG-2 & 79.43 & 55.96 - 81.51 & 95.17 & 85.96 - 95.09 & 94.67 & 86.11 - 93.89 & \\
\quad AI-TG-2 & 84.96 & 73.71 - 96.09 & 96.79 & 93.84 - 99.42 & 96.22 & 93.72 - 98.69 & 0.21 \\

\textbf{Examination-3} & & & & & & & \\
\quad T-TG-1 & 59.32 & 60.06 - 69.50 & 88.02 & 83.08 - 92.95 & 88.11 & 83.88 - 92.35 & \\
\quad AI-TG-1 & 71.76 & 69.10 - 94.80 & 90.35 & 92.15 - 98.57 & 89.33 & 91.20 - 97.35 & 0.67 \\
\quad T-TG-2 & 83.35 & 57.06 - 85.40 & 95.46 & 85.82 - 94.70 & 94.28 & 85.26 - 93.41 & \\
\quad AI-TG-2 & 88.05 & 75.63 - 98.78 & 97.77 & 95.64 - 99.59 & 97.39 & 95.38 - 99.27 & 0.03 \\

\textbf{Examination-4} & & & & & & & \\
\quad T-TG-1 & 63.30 & 52.52 - 74.72 & 88.77 & 84.45 - 93.08 & 88.95 & 84.94 - 92.94 & \\
\quad AI-TG-1 & 68.96 & 70.86 - 95.31 & 90.02 & 93.93 - 99.59 & 88.95 & 93.70 - 98.84 & 0.83 \\
\quad T-TG-2 & 83.46 & 54.86 - 83.27 & 96.92 & 85.13 - 94.86 & 96.28 & 84.92 - 92.97 & \\
\quad AI-TG-2 & 90.31 & 80.83 - 98.04 & 97.79 & 95.62 - 99.59 & 97.78 & 95.95 - 99.50 & 0.10 \\
\hline
 & & & \multicolumn{2}{c}{\textbf{100 random cases}}  & & & \\
\textbf{Examination-1} & & & & & & & \\
\quad T-TG-1 & 63.63 & 53.16 - 75.82 & 90.85 & 83.58 - 97.58 & 90.67 & 85.07 - 92.26 &  \\
\quad AI-TG-1 & 64.94 & 59.34 - 90.92 & 91.30 & 91.45 - 99.80 & 90.67 & 91.38 - 99.02 & 1.00 \\
\quad T-TG-2 & 74.92 & 49.52 - 82.25 & 96.07 & 85.06 - 97.15 & 95.34 & 85.40 - 95.82 &  \\
\quad AI-TG-2 & 78.29 & 60.67 - 92.04 & 96.56 & 92.34 - 99.64 & 96.34 & 92.96 - 99.36 & 0.46 \\

\textbf{Examination-2} & & & & & & & \\
\quad T-TG-1 & 60.97 & 43.86 - 77.96 & 85.67 & 77.67 - 93.53 & 86.67 & 82.70 - 94.59 &  \\
\quad AI-TG-1 & 75.96 & 56.85 - 93.8 & 91.33 & 84.64 - 97.84 & 92.67 & 87.82 - 97.38 & 0.18 \\
\quad T-TG-2 & 83.83 & 65.42 - 98.58 & 95.80 & 91.06 - 99.70 & 96.22 & 92.78 - 99.37 &  \\
\quad AI-TG-2 & 86.5 & 70.40 - 98.68 & 97.91 & 94.90 - 99.92 & 97.78 & 95.33 - 99.7 & 0.17 \\

\textbf{Examination-3} & & & & & & & \\
\quad T-TG-1 & 62.17 & 47.02 - 79.20 & 88.07 & 80.93 - 94.91 & 90.78 & 85.41 - 96.07 &  \\
\quad AI-TG-1 & 64.12 & 63.46 - 86.52 & 92.05 & 91.88 - 99.82 & 91.56 & 92.20 - 99.30 & 0.92 \\
\quad T-TG-2 & 76.09 & 49.35 - 80.66 & 96.37 & 85.71 - 97.78 & 95.89 & 86.50 - 96.48 &  \\
\quad AI-TG-2 & 80.62 & 66.85 - 92.97 & 97.48 & 93.86 - 99.99 & 97.34 & 94.38 - 99.83 & 0.26 \\

\textbf{Examination-4} & & & & & & & \\
\quad T-TG-1 & 59.58 & 45.88 - 77.38 & 87.96 & 80.43 - 92.25 & 89.89 & 84.31 - 95.41 &  \\
\quad AI-TG-1 & 61.12 & 62.46 - 91.15 & 89.01 & 90.67 - 99.79 & 88.78 & 91.98 - 99.37 & 0.92 \\
\quad T-TG-2 & 75.98 & 47.47 - 75.53 & 95.56 & 81.77 - 95.79 & 95.78 & 83.14 - 94.32 &  \\
\quad AI-TG-2 & 79.62 & 69.83 - 92.14 & 96.68 & 92.64 - 99.92 & 97.00 & 93.79 - 99.91 & 0.07 \\
\bottomrule
\end{tabular}
\end{table}

\begin{figure}[h]
    \centering
    \includegraphics[width=1\textwidth,height=\textheight,keepaspectratio]{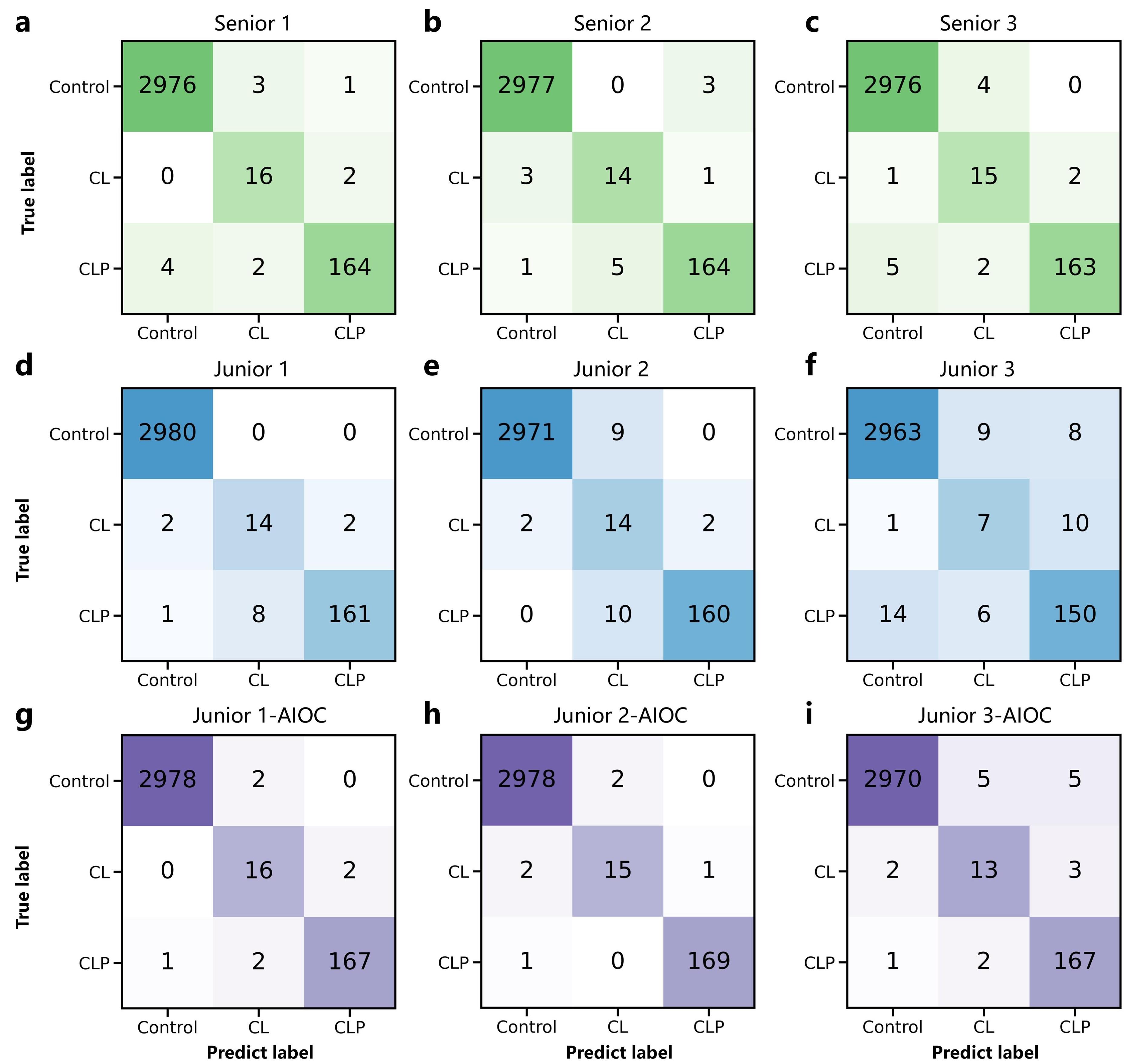}
    \caption{\textbf{Confusion matrix of senior radiologists, junior radiologists, and AIOC-assisted junior radiologists for the classification of three types of orofacial clefts: control, cleft lip (CL), and cleft lip and palate (CLP).} a-c. Performance of senior R1, R2 and R3. d-f. Performance of junior R4, R5 and R6. g-i. Performance of AIOC-assisted junior radiologists R4, R5 and R6. CL: cleft lip; CLP: cleft lip and palate.}
    \label{fig: S2}
\end{figure}

\begin{sidewaystable}[!h]
  \centering
  \caption{Characteristics of the three datasets used in this study. The table presents the distribution of control, CL, and CLP cases and images across all datasets, along with maternal and gestational age information, and ultrasound equipment specifications. All data represent case numbers and corresponding image counts, with percentages shown in parentheses for the OC-6000 subsets.}
    \begin{tabular}{@{}lcccccc@{}}
    
    \toprule
    \multirow{2}[4]{*}{Groups} & \multicolumn{3}{c}{OC-6000} & \multicolumn{1}{c}{\multirow{2}[4]{*}{OC-GT3000}} & \multicolumn{1}{c}{\multirow{2}[4]{*}{OC-Early}} & \multicolumn{1}{c}{\multirow{2}[4]{*}{Total}} \\
\cmidrule{2-4}    \multicolumn{1}{c}{} & \multicolumn{1}{c}{Training (80\%)} & \multicolumn{1}{c}{Validation (10\%)} & \multicolumn{1}{c}{Testing (10\%)} &       &       &  \\
    \midrule
    Control (Case/Image) & \multicolumn{1}{c}{4168/18762 (84.8\%)} & \multicolumn{1}{c}{415/1867 (81.7\%)} & \multicolumn{1}{c}{495/2228 (84.5\%)} & \multicolumn{1}{c}{2980/14672} & \multicolumn{1}{c}{18/51} & \multicolumn{1}{c}{8076/37580} \\
    CL (Case/Image) & \multicolumn{1}{c}{64/415 (1.3\%)} & \multicolumn{1}{c}{18/117 (3.5\%)} & \multicolumn{1}{c}{21/136 (3.6\%)} & \multicolumn{1}{c}{18/128} & \multicolumn{1}{c}{1/13} & \multicolumn{1}{c}{122/809} \\
    CLP (Case/Image) & \multicolumn{1}{c}{684/4513 (13.9\%)} & \multicolumn{1}{c}{75/494 (14.8\%)} & \multicolumn{1}{c}{70/462 (11.9\%)} & \multicolumn{1}{c}{170/1048} & \multicolumn{1}{c}{18/233} & \multicolumn{1}{c}{1017/6750} \\
    \midrule
    Total (Case/Image) & \multicolumn{1}{c}{4916/23690} & \multicolumn{1}{c}{508/2478} & \multicolumn{1}{c}{586/2826} & \multicolumn{1}{c}{3168/15848} & \multicolumn{1}{c}{37/297} & \multicolumn{1}{c}{9215/45139} \\
    \midrule
    \midrule
    \textbf{Age} \\
    Maternal Age (years) & \multicolumn{1}{c}{29.43±6.71} & \multicolumn{1}{c}{28.56±4.35} & \multicolumn{1}{c}{29.12±5.99} & \multicolumn{1}{c}{29.56±6.17} & \multicolumn{1}{c}{27.55±4.03} & \multicolumn{1}{c}{29.34±6.81} \\
    Gestational Age (weeks) & \multicolumn{1}{c}{22.84±1.99} & \multicolumn{1}{c}{22.54±3.21} & \multicolumn{1}{c}{22.62±3.89} & \multicolumn{1}{c}{22.46±1.44} & \multicolumn{1}{c}{16.59±0.72} & \multicolumn{1}{c}{22.51±2.15} \\
    \midrule
    \textbf{Fetal gender (Cases)} & & & \\
    Male & 2,409 & 302& 315 & 1,601 & 17 & 4,664 \\
    Female & 2,460 & 199& 261& 1,532 & 20 & 4,472\\
    Ambiguous & 47 & 7 & 10& 35 & 0 & 99\\
    \midrule
    \textbf{Cleft type (Cases)} & & & \\
    Unilateral & 665 & 85 & 73 & 148 & 18 & 989\\
    Bilateral & 71 & 7& 16 & 24 & 1 & 119\\
    Medline & 12 & 1 & 2 & 16 & 0 & 31\\
    \midrule
    \textbf{Associated structural anomaly (Cases)} & & & \\
    Central nervous system & 6 & 2& 1& 4 & 0 & 13 \\
    Cardiovascular system & 8 & 1& 1& 4 & 2 & 16 \\
    Musculoskeletal system & 7 & 1 & 0& 1 & 0 & 9\\
    Genitourinary system & 3 & 0 & 1& 1 & 1 & 6 \\
    Multiple structural anomaly & 23 & 3& 5& 8 & 1 & 40 \\
    \midrule
    \textbf{Ultrasound Equipment (Cases)} \\
    GE E6 & 381   & 46    & 54    & 290   & 0     & 771 \\
    GE E8 & 51    & 5     & 6     & 33    & 1     & 96 \\
    GE E10 & 3990  & 412   & 475   & 2571  & 30    & 7478 \\
    Philips EPIQ7 & 91    & 9     & 11    & 59    & 4     & 174 \\
    Philips Affiniti70 & 188   & 19    & 22    & 121   & 0     & 350 \\
    Philips IU22 & 23    & 2     & 3     & 15    & 0     & 43 \\
    Samsung HERA W10 & 7     & 1     & 1     & 4     & 1     & 14 \\
    Mindray Resona 8  & 185   & 14    & 14    & 75    & 1     & 289 \\
    \bottomrule
    \end{tabular}%
  \label{tab: S1}%
\end{sidewaystable}

\clearpage
\newpage
\section*{Supplementary Materials}

\maketitle 
\setcounter{table}{0}  
\setcounter{figure}{0}
\setcounter{section}{0}
\setcounter{equation}{0}
\renewcommand{\thetable}{S\arabic{table}}
\renewcommand{\thefigure}{S\arabic{figure}}
\renewcommand{\thesection}{S\arabic{section}}
\renewcommand{\theequation}{S\arabic{equation}}

\begin{table}[h]
\centering
\caption{List of 22 hospitals from our multicenter study and the amount of data provided for fetal OC and healthy cases. The total data includes both the included experimental data and the excluded data.}
\label{tab: S2}
\begin{tabular}{lcc}
\hline
Hospital & OC cases & healthy cases \\
\hline
Changshu Hospital of Traditional Chinese Medicine & 51 & 1601 \\
Changzhou Maternal and Child Health Care Hospital & 41 & 285 \\
Hangzhou Women's Hospital & 100 & 0 \\
Jiangsu Taizhou People's Hospital & 82 & 156 \\
Nanjing Women and Children's Healthcare Hospital & 52 & 0 \\
Northwest Women's and Children's Hospital & 183 & 907 \\
Qilu Hospital of Shandong University & 148 & 143 \\
Roicare Hospital \& Clinics & 67 & 163 \\
Second Hospital of Anhui Medical University & 46 & 1197 \\
Shanxi Children's Hospital & 110 & 96 \\
Shenzhen Guangming District People's Hospital & 11 & 0 \\
Shenzhen Luohu Hospital & 142 & 1234 \\
Sichuan Provincial Maternity and Child Health Care Hospital & 5 & 20 \\
Suzhou Municipal Hospital & 103 & 56 \\
The Affiliated Hospital of Nanjing University Medical School & 16 & 72 \\
The Affiliated Hospital of Xuzhou Medical University & 92 & 2312 \\
The First Affiliated Hospital of Nanchang University & 69 & 1484 \\
The First Hospital of China Medical University & 47 & 60 \\
The Fourth Affiliated Hospital of Soochow University & 3 & 1482 \\
The People's Hospital of Guangxi Zhuang Autonomous Region & 18 & 49 \\
The Second Hospital of Dalian Medical University & 6 & 262 \\
Zhejiang Provincial People's Hospital & 33 & 153 \\
\hline
\end{tabular}
\end{table}

\begin{table}[h]
\centering
\caption{View classification performance of the AIOC system on the OC-6000 and OC-GT3000 datasets was evaluated across four key views: Normal Lip View (NLV), Normal Alveolus and Palate View (NAPV), Cleft Lip View (CLV), and Cleft Alveolus and Palate View (CAPV). Data are presented as \%. AUC: area under the receiver operating characteristic curve; CI: confidence interval; FPR: false positive rate; FNR: false negative rate; F1: F1-score.}
\label{tab: S3}
\begin{tabular}{lcccccccc}
\hline
\textbf{Characteristic} & AUC (\%) & Sensitivity (\%) & Specificity (\%) & Accuracy & FNR & FPR & F1 & Youden \\
 & (95\%CI) & (95\%CI) & (95\%CI) &  (\%) & (\%) & (\%) & (\%) & Index \\
\hline
\textbf{OC-6000} & & & & & & & & \\
\quad NAPV & 99.49 & 98.2 & 99.92 & 99.27 & 1.8 & 0.51 & 97.9 & 0.98 \\
 & (99.85 - 99.98) & (97.03 - 99.36) & (99.21 - 99.78) & & & & & \\
\quad NLV & 98.94 & 99.13 & 99.84 & 99.06 & 0.87 & 1.06 & 99.26 & 0.98 \\
 & (99.67 - 99.95) & (98.7 - 99.55) & (98.32 - 99.56) & & & & & \\
\quad CLV & 99.25 & 96.99 & 99.7 & 98.99 & 3.01 & 0.75 & 95.69 & 0.96 \\
 & (99.41 - 99.79) & (95.15 - 98.83) & (98.92 - 99.59) & & & & & \\
\quad CAPV & 99.74 & 93.3 & 99.53 & 99.27 & 6.7 & 0.26 & 94.89 & 0.93 \\
 & (98.8 - 99.94) & (89.91 - 96.69) & (99.54 - 99.93) & & & & & \\
\hline
\textbf{OC-GT3000} & & & & & & & & \\
\quad NAPV & 99.69 & 99.38 & 99.96 & 99.59 & 0.62 & 0.31 & 99.37 & 0.99 \\
 & (99.92 - 99.99) & (99.16 - 99.59) & (99.59 - 99.8) & & & & & \\
\quad NLV & 99.54 & 99.63 & 99.97 & 99.59 & 0.37 & 0.46 & 99.66 & 0.99 \\
 & (99.94 - 99.99) & (99.5 - 99.75) & (99.37 - 99.7) & & & & & \\
\quad CLV & 99.79 & 95.5 & 99.83 & 99.58 & 4.5 & 0.26 & 94.55 & 0.95 \\
 & (99.7 - 99.93) & (93.84 - 97.16) & (99.66 - 99.82) & & & & & \\
\quad CAPV & 99.72 & 92.56 & 99.79 & 99.56 & 7.44 & 0.21 & 93.11 & 0.92 \\
 & (99.72 - 99.86) & (90.29 - 94.84) & (99.72 - 99.86) & & & & & \\
\hline
\end{tabular}
\end{table}

\begin{table}[!t]
\centering
\caption{\textbf{Structure detection performance of the AIOC system on the OC-6000 and OC-GT3000 datasets.} mAP: mean Average Precision.}
\label{tab: S4}
\begin{tabular}{p{4cm}p{2cm}p{3cm}p{2cm}p{2cm}p{2cm}}
\hline
\textbf{Structure} &\textbf{ upper lip} & \textbf{alveolar ridge} & \textbf{cleft lip} & \textbf{cleft alveolus} & \textbf{cleft palate} \\
\hline
\textbf{mAP of OC-6000} & 0.93 & 0.73 & 0.86 & 0.84 & 0.44 \\
\textbf{mAP of OC-GT3000} & 0.99 & 0.97 & 0.80 & 0.77 & 0.42 \\
\hline
\end{tabular}
\end{table}

\begin{table}[h]
    \centering
    \caption{Comparative performance based on F1-score for fetal OC case-based diagnosis across different gestational weeks (18–28 weeks), with AIOC compared to senior, junior, and junior-AIOC radiologists. Due to the small number of cases in the 18-20 and 25-28 week groups, these weeks were combined for statistical analysis. GA: gestational age; SD: standard deviation.}
    \label{tab: S6-Gestational week}
    \begin{tabular}{lccccccc}
        \hline
        GA/ Groups & 18 - 20 weeks & 21 weeks & 22 weeks & 23 weeks & 24 weeks & 25 - 28 weeks & SD \\
        \hline
        AIOC & 100.00 & 99.80 & 82.40 & 93.91 & 94.32 & 93.96 & 5.84 \\
        Seniors & 82.22 & 99.93 & 100 & 95.79 & 92.99 & 100.00 & 6.35 \\
        Juniors & 59.37 & 99.93 & 82.47 & 85.31 & 84.95 & 97.04 & 13.11 \\
        \hline
    \end{tabular}
\end{table}

\begin{table}[h]
\centering
\caption{Time requirement of the AIOC, senior radiologists, junior radiologists, and AIOC-assisted junior radiologists for case-based diagnosis on OC-GT3000 dataset. Diagnostic time was compared using a two-sided Welch’s t-test}.
\label{tab: S6-Time requirement}
\begin{tabular}{lcccc}
\toprule
& \multicolumn{2}{c}{Time requirement} & \multirow{2}{*}{$t$-statistic} & \multirow{2}{*}{$p$-value} \\
\cmidrule(r){2-3}
 & Single case (Seconds) & Total cases (Hours) & & \\
\midrule
AIOC & 0.32 & 0.28 & - & - \\
Senior Average & 10.54 & 9.34 & 7.40 & $9.7 \times 10^{-14}$ \\
Junior Average & 11.93 & 10.57 & 9.74 & $7.0\times 10^{-8}$ \\
Junior-AIOC Average & 5.31 & 4.67 & 12.69 & $3.5 \times 10^{-166}$ \\
\bottomrule
\end{tabular}
\end{table}

\begin{table}[h]
\centering
\caption{Automation Bias Analysis in AI-Assisted Diagnosis}
\begin{tabular}{lccccc}
\hline
\multirow{2}{*}{Doctor} & \multirow{2}{*}{AI Accuracy} & \multicolumn{2}{c}{Doctor Accuracy} & \multicolumn{2}{c}{Automation Bias Metrics} \\
\cmidrule(lr){3-4} \cmidrule(lr){5-6}
 & & Without AI & With AI & Over-reliance & Appropriate Reliance \\
\hline
Junior R4 & 99.5\% & 99.6\% & 99.8\% & 11.8\% & 0.4\% \\
Junior R5 & 99.5\% & 99.3\% & 99.8\% & 11.8\% & 0.7\% \\
Junior R6 & 99.5\% & 98.5\% & 99.4\% & 5.9\% & 1.2\% \\
\hline
\end{tabular}
\label{tab:automation_bias}
\end{table}

\begin{table}[h]
  \centering
  \caption{Detailed results of four difficult cases out of 100 random cases in medical training, diagnosed by traditional (T-TG) and AI-augmented training groups (AI-TG).}
    \begin{tabular}{ccccc}
    \toprule
          & \multirow{2}[2]{*}{Type} & \multirow{2}[2]{*}{Correct Num.} & \multicolumn{2}{c}{Correct Proportion} \\
          &       &       & T-TG  & AI-TG \\
    \midrule
    Examination-1 & CL    & 8     & 37.50\% & 62.50\% \\
    Examination-2 & CLP   & 7     & 28.57\% & 71.43\% \\
    Examination-3 & CLP   & 12    & 41.67\% & 58.33\% \\
    Examination-4 & CL    & 5     & 40.00\% & 60.00\% \\
    \bottomrule
    \end{tabular}%
  \label{tab:S8}%
\end{table}%

\begin{figure}[h]
    \centering
    \includegraphics[width=1\textwidth,height=\textheight,keepaspectratio]{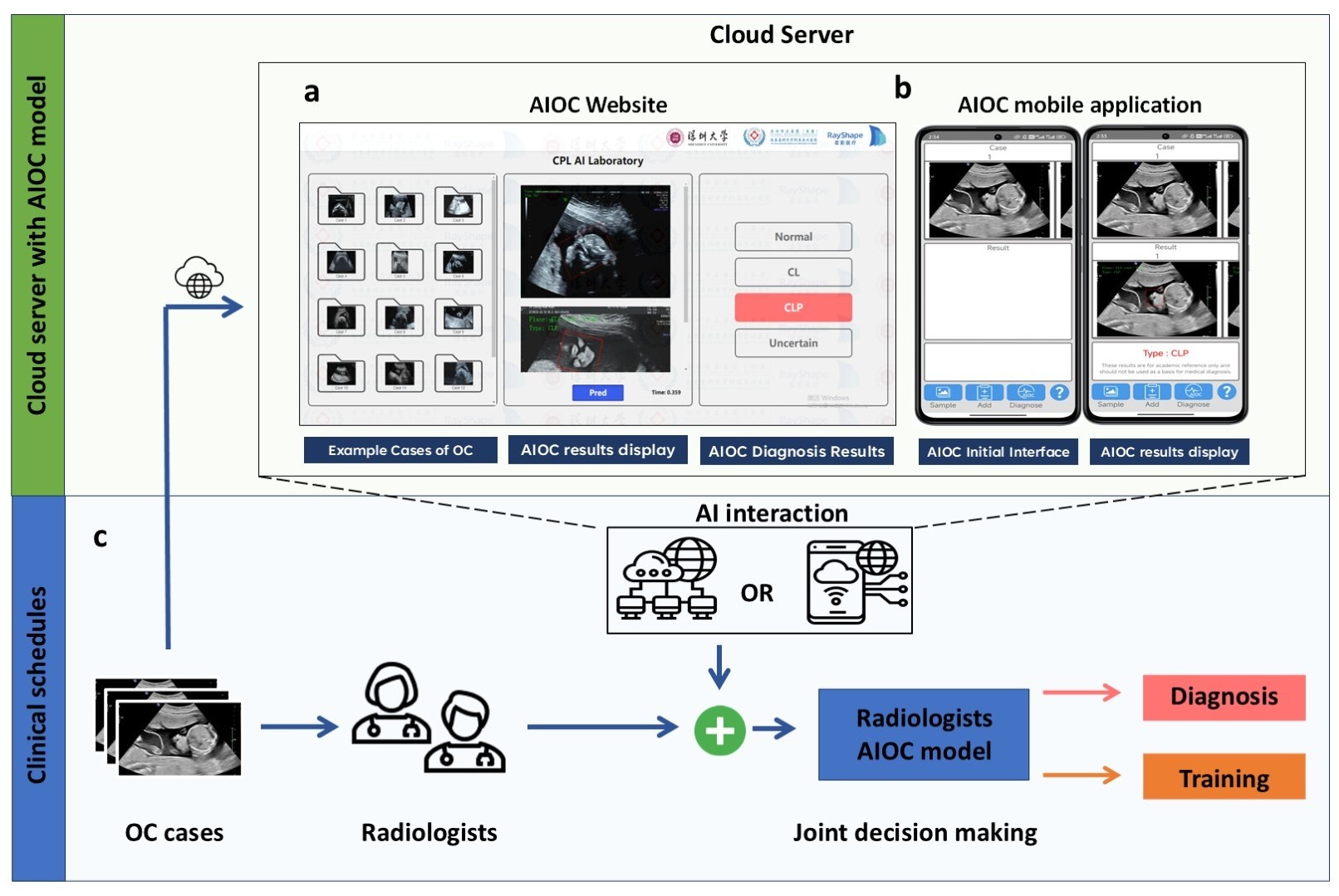}
    \caption{\textbf{Application of the AIOC system in orofacial case (OC) diagnosis and result display.}
(a) Web Interface – Includes interfaces for example cases and uploading new cases, predicting views and structures of each OC case image, and the final diagnosis prediction interface.
(b) Mobile Application – Displays the initial case upload screen and interfaces for predicting views, structures, and diagnostic outcomes of OC cases.
(c) Fetal OC and Healthy Cases – The workflow for using the AIOC system includes radiologists uploading OC and healthy cases to the AI interaction interface on the website or mobile application. Through joint decision-making with AIOC, the system supports diagnostic assistance and training purposes.}
    \label{fig: S3}
\end{figure}

\begin{figure}[t]
    \centering
    \includegraphics[width=1\textwidth,height=\textheight,keepaspectratio]{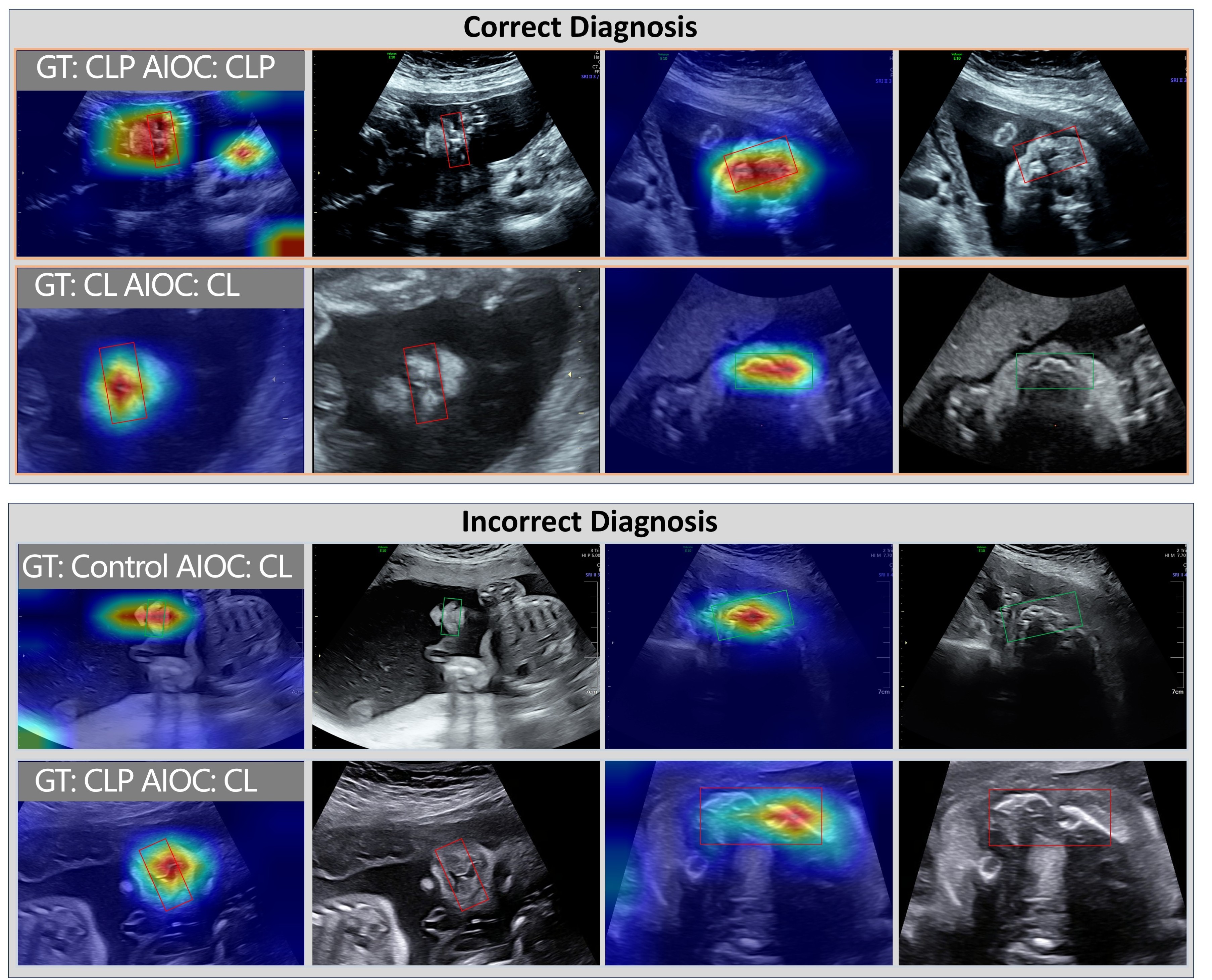}
    \caption{\textbf{Grad-CAM visualizations for both correctly and incorrectly diagnosed cases.} Each row displays two different ultrasound views per case. For each view, the left image shows Grad-CAM heatmap visualization highlighting regions of model attention (warmer colors indicate higher attention), and the right image shows the corresponding ultrasound with bounding boxes overlaid on anatomical structures. Top panel (Correct Diagnosis): Two cases where the model correctly identified CL and CLP. Bottom panel (Incorrect Diagnosis): Two cases demonstrating diagnostic errors, where ground truth labels differ from model predictions (GT: Control vs AIOC: CL in the first case; GT: CLP vs AIOC: CL in the second case).}
    \label{fig: correct_incorrect}
\end{figure}

\end{document}